\documentclass[letterpaper]{article} 
\usepackage{aaai2026}  
\usepackage{times}  
\usepackage{helvet}  
\usepackage{courier}  
\usepackage[hyphens]{url}  
\usepackage{graphicx} 
\usepackage{amsmath}
\usepackage{amsfonts}
\usepackage{booktabs}
\usepackage{natbib}  
\usepackage{caption} 
\usepackage{algorithm}
\usepackage{algorithmic}

\usepackage{stfloats}
\usepackage{newfloat}
\usepackage{listings}
\DeclareCaptionStyle{ruled}{labelfont=normalfont,labelsep=colon,strut=off} 
\floatstyle{ruled}
\newfloat{listing}{tb}{lst}{}
\floatname{listing}{Listing}
\usepackage{hyperref}
\title{MedRep: Medical Concept Representations for General Electronic Health Record Foundation Models}
\author{
    Junmo Kim\textsuperscript{\rm 1}, Namkyeong Lee\textsuperscript{\rm 2}, Jiwon Kim\textsuperscript{\rm 3}, Kwangsoo Kim\textsuperscript{\rm 4,5,6}
}
\affiliations{
    \textsuperscript{\rm 1}Interdisciplinary Program in Bioengineering, Seoul National University\\
    \textsuperscript{\rm 2}Dept. of Industrial and Systems Engineering, KAIST \\
    \textsuperscript{\rm 3}Interdisciplinary Program of Medical Informatics, Seoul National University \\
    \textsuperscript{\rm 4}Dept. of Transdisciplinary Medicine, ICMIT, Seoul National University Hospital \\
    \textsuperscript{\rm 5}Center for Data Science, Healthcare AI Research Institute, Seoul National University Hospital \\
    \textsuperscript{\rm 6}Dept. of Medicine, College of Medicine, Seoul National University\\

}

\usepackage{bibentry}

\begin{document}

\maketitle

\begin{abstract}
Electronic health record (EHR) foundation models have been an area ripe for exploration with their improved performance in various medical tasks. Despite the rapid advances, there exists a fundamental limitation: Processing unseen medical codes out of vocabulary. This problem limits the generalizability of EHR foundation models and the integration of models trained with different vocabularies. To alleviate this problem, we propose a set of novel medical concept representations (MedRep) for EHR foundation models based on the observational medical outcome partnership (OMOP) common data model (CDM). For concept representation learning, we enrich the information of each concept with a minimal definition through large language model (LLM) prompts and complement the text-based representations through the graph ontology of OMOP vocabulary. Our approach outperforms the vanilla EHR foundation model and the model with a previously introduced medical code tokenizer in diverse prediction tasks. We also demonstrate the generalizability of MedRep through external validation. Our code implementation is publicly available at \url{https://github.com/kicarussays/MedRep}
\end{abstract}


\section{Introduction}

Electronic health records (EHRs) store a lot of data generated from a hospital, such as diagnoses, measurements, prescriptions, and procedures. Most tertiary hospitals in many countries have adopted EHRs to manage hospital data \citep{parasrampuria2019hospitals, liang2021adoption}. With the widespread use of EHRs, numerous studies have utilized EHR data and machine learning techniques for tasks like medical event prediction \citep{huang2023development}, drug recommendation \citep{zhang2023enhancing}, and patient monitoring \citep{kim2024deep}. With the availability of large patient data and the tremendous success of language models, developing EHR foundation models based on medical code and language models has become an area ripe for exploration \citep{li2020behrt, rasmy2021med, yang2023transformehr, renc2024zero}. The medical history of each patient can be represented as a sequence of medical codes, which is called a patient trajectory \citep{bornet2025comparing}. Each code and trajectory corresponds to a word and a sentence in natural language. Along with similar problem settings, EHR foundation models have been trained to comprehend medical context by paradigms of language models and utilized to predict several medical events, such as in-hospital mortality, long length of stay, readmission, and various diseases \citep{li2020behrt, guo2024multi}. 

Despite the rapid advances, EHR foundation models have a fundamental limitation: The discrepancy of medical code vocabularies from different institutions. For example, suppose two hospitals store diagnosis records using different vocabularies SNOMED-CT \citep{donnelly2006snomed} and ICD-10 \citep{quan2005coding}, respectively. In this case, a model trained from one hospital may not operate at the other hospital because the unseen medical codes might be treated as unknown codes. To mitigate the difference in data, a lot of medical institutions have introduced an observational medical outcome partnership (OMOP) common data model (CDM), developed and maintained by the Observational Health Data Sciences and Informatics (OHDSI) initiative \citep{stang2010advancing}. This model enables the transformation of different EHR databases into a standardized format. Instead of directly relying on various medical vocabularies, such as SNOMED-CT \citep{donnelly2006snomed}, ICD-10 \citep{quan2005coding}, RxNorm \citep{liu2005rxnorm}, and LOINC \citep{mcdonald2003loinc}, the OMOP CDM standardizes all those medical codes under a single unified vocabulary: the OMOP vocabulary. More than 9 million medical codes from dozens of medical vocabularies can be systemically mapped to standardized OMOP concept IDs. As a result, EHR foundation models built on OMOP CDM can seamlessly operate across any institution adopting the OMOP standard.

However, merging OMOP CDM-based EHR data across institutions is still hardly available due to patient privacy regulations. Thus, EHR foundation models built on OMOP CDM must be trained separately at each institution, even though they share a common data structure and vocabulary. Additionally, hospitals using OMOP CDM may use different concept IDs for the same concept \citep{kim2025pretrained}. For instance, to map the prescription of aspirin 100MG, there exist diverse relevant concept IDs, such as 1113143 (aspirin 100MG Oral Tablet), 42483115 (Aspirin 100 MG Oral Capsule), 40012940 (Aspirin 100MG Oral Solution), 35136978 (Aspirin 100MG Oral Tablet by Pfizer), and 36893848 (Aspirin 100MG/ML Oral Solution). While these concepts are similar, EHR foundation models may treat them differently because they are associated with distinct concept IDs and tokenized indices. As a result, when a model is transferred to other institutions, it may fail to recognize semantically similar but previously unseen concepts, limiting its ability to operate across external datasets generally.

In this study, we propose MedRep, a set of OMOP medical concept representations that can be directly utilized across any EHR foundation model. MedRep can replace traditional learnable token embeddings, preventing EHR foundation models from treating the embeddings of the same tokens differently. For MedRep training, we first generate brief descriptions of the clinical background and context for each concept using large language model (LLM) prompts \citep{schick2020exploiting}. Next, we learn the generated descriptions through a masked language model (MLM) \citep{devlin2019bert} to produce text-based representations. These representations are then enhanced through graph representation learning and the graph ontology of the OMOP vocabulary. Through MedRep, we enhance the compatibility of EHR foundation models, enhancing downstream task performance in both internal and external validation. There exists concurrent work, MedTok \citep{su2025multimodal}; however, it has not been explored to externally validate the performance using separate datasets. We compare the internal and external validation performance of EHR foundation models with no pretrained representations, with MedTok, and with MedRep.

\section{Results}
\paragraph{Data and target outcomes} We used two publicly available EHR datasets: MIMIC-IV (version 2.2) \citep{johnson2023mimic} and EHRSHOT \citep{wornow2023ehrshot}. To assess the compatibility of MedRep, we first pretrained and finetuned EHR foundation models on MIMIC-IV, then evaluated their predictive performance using the MIMIC-IV hold-out test set (internal validation) and the full EHRSHOT dataset (external validation), without any further parameter updates. Since EHRSHOT is originally formatted in OMOP CDM while MIMIC-IV is not, we converted the original MIMIC-IV into OMOP format using the official GitHub repository of OHDSI (\url{https://github.com/OHDSI/MIMIC}). We used four OMOP CDM tables, including condition occurrence, drug exposure, measurement, and procedure occurrence. For pretraining, we used 85\% of the MIMIC-IV patients (70\% for training and 15\% for validation). The remaining 15\% of patients were used for finetuning, which was further split into training, validation, and test (internal validation) sets at a 6:2:2 ratio. Since EHR foundation models rely on longitudinal clinical records, we included only patients who had been hospitalized for more than two days \citep{kim2025pretrained}. Baseline characteristics of the datasets are summarized in Table \ref{tab:characteristics}. Both datasets contained approximately 26,000 medical concepts, and MIMIC-IV had a higher proportion of male patients compared to EHRSHOT (62.87\% versus 54.60\%, respectively).

\begin{table}[!]
  \caption{Baseline characteristics of datasets.}
\resizebox{\columnwidth}{!}{
\begin{tabular}[!]{lll}
\toprule
 & \textbf{MIMIC-IV} & \textbf{EHRSHOT} \\
 \midrule
Patients, n & 257,992 & 4,403 \\
Records, n & 1,114,054,364 & 2,942,281 \\
Visits, n & 2,468,013 & 116,060 \\
Concept IDs, n & 26,894 & 26,303 \\
Age, mean (std) & 52.54 (20.89) & 52.33 (17.31) \\
Male sex, n & 62.87\% & 54.6\% \\
 \midrule
Clinical outcomes, \% (case / total) &  &  \\
In-hospital mortality & 3.68\% (239 / 6499) & 6.57\% (289 / 4402) \\
Long length of stay & 30.90\% (2008 / 6499) & 27.01\% (1189 / 4402) \\
Readmission & 2.02\% (131 / 6499) & 9.45\% (416 / 4402) \\
 \midrule
Phenotypes, \% (case / total) &  &  \\
Acute and unspecified renal failure & 10.41\% (1280 / 12301) & 7.27\% (191 / 2628) \\
Acute cerebrovascular disease & 4.38\% (539 / 12301) & 5.29\% (139 / 2628) \\
Acute myocardial infarction & 2.16\% (266 / 12301) & 2.97\% (78 / 2628) \\
Cardiac dysrhythmias & 14.08\% (1732 / 12301) & 18.42\% (484 / 2628) \\
Chronic kidney disease & 7.30\% (898 / 12301) & 7.80\% (205 / 2628) \\
Chronic obstructive pulmonary disease & 5.44\% (669 / 12301) & 5.63\% (148 / 2628) \\
Complications of surgical/medical care & 53.56\% (6589 / 12301) & 58.71\% (1543 / 2628) \\
Conduction disorders & 3.72\% (458 / 12301) & 7.80\% (205 / 2628) \\
Congestive heart failure; nonhypertensive & 7.39\% (909 / 12301) & 7.08\% (186 / 2628) \\
Coronary atherosclerosis and related & 10.85\% (1335 / 12301) & 10.01\% (263 / 2628) \\
Diabetes mellitus with complications & 11.61\% (1428 / 12301) & 11.30\% (297 / 2628) \\
Diabetes mellitus without complication & 63.21\% (7775 / 12301) & 37.06\% (974 / 2628) \\
Disorders of lipid metabolism & 19.34\% (2379 / 12301) & 17.16\% (451 / 2628) \\
Essential hypertension & 22.30\% (2743 / 12301) & 20.47\% (538 / 2628) \\
Fluid and electrolyte disorders & 12.63\% (1554 / 12301) & 10.84\% (285 / 2628) \\
Gastrointestinal hemorrhage & 2.72\% (335 / 12301) & 2.93\% (77 / 2628) \\
Hypertension with complications & 11.32\% (1393 / 12301) & 12.18\% (320 / 2628) \\
Other liver diseases & 25.28\% (3110 / 12301) & 37.48\% (985 / 2628) \\
Other lower respiratory disease & 28.02\% (3447 / 12301) & 36.95\% (971 / 2628) \\
Other upper respiratory disease & 35.50\% (4367 / 12301) & 39.92\% (1049 / 2628) \\
Pleurisy; pneumothorax; pulmonary collapse & 3.87\% (476 / 12301) & 9.32\% (245 / 2628) \\
Pneumonia & 3.63\% (446 / 12301) & 3.23\% (85 / 2628) \\
Respiratory failure; insufficiency; arrest & 3.93\% (484 / 12301) & 3.54\% (93 / 2628) \\
Septicemia (except in labor) & 4.05\% (498 / 12301) & 3.69\% (97 / 2628) \\
 \midrule
In-hospital events &  &  \\
Urinary tract infection & 11.88\% (2127 / 17906) & 12.88\% (482 / 3741) \\
Fracture & 5.81\% (1028 / 17680) & 7.63\% (290 / 3802) \\
Sepsis & 8.85\% (1599 / 18070) & 18.28\% (726 / 3972) \\
Pneumonia & 11.80\% (2110 / 17879) & 19.34\% (749 / 3872) \\
Myocardial infarction & 2.80\% (511 / 18258) & 10.06\% (393 / 3907) \\
\bottomrule
\end{tabular}}
  \label{tab:characteristics}
\end{table}

\begin{figure}[!t]
  \centerline{\includegraphics[width=\columnwidth]{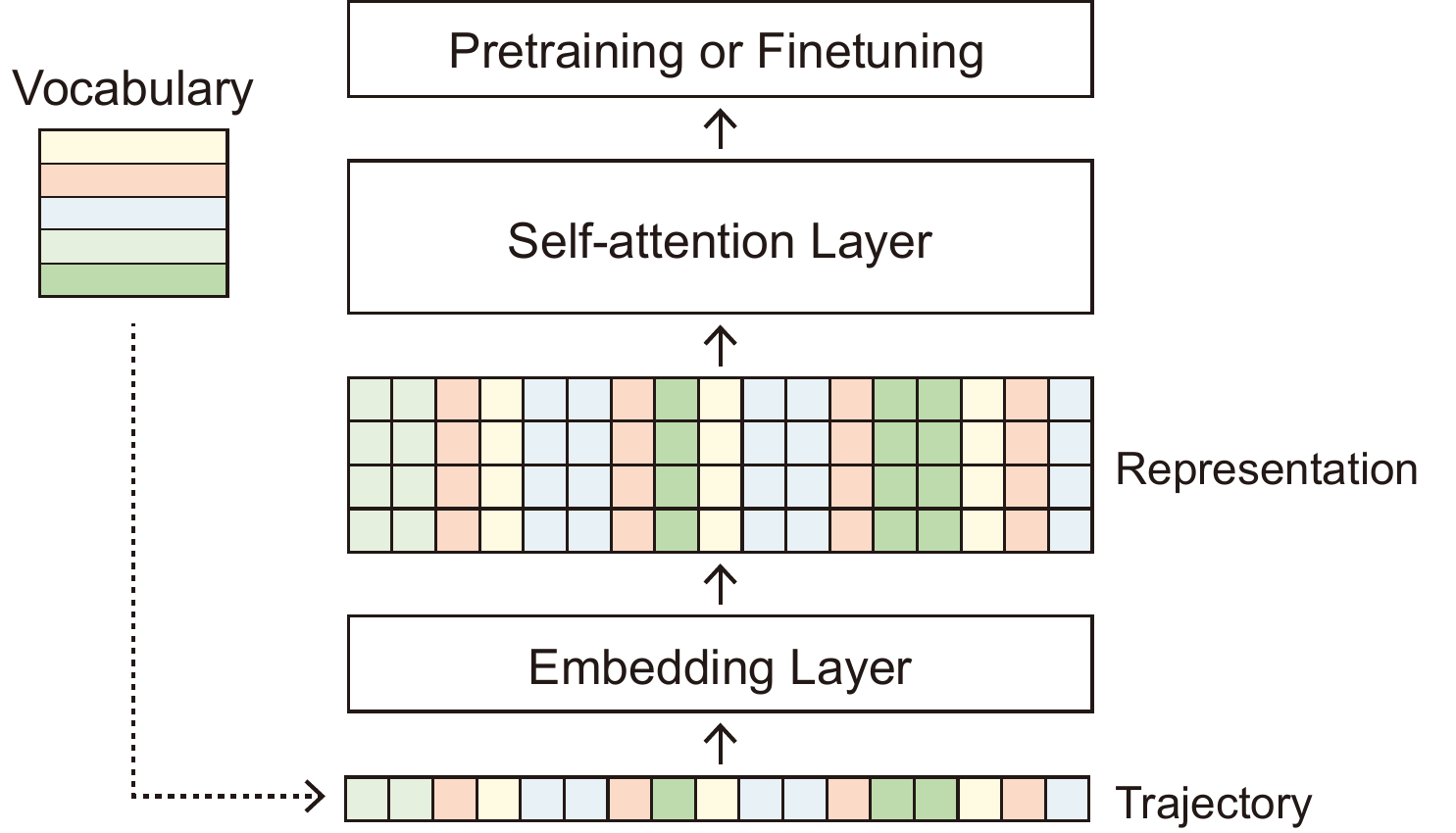}}
  \caption{Basic structure of electronic health record foundation models. A patient trajectory is a sequence of medical code indices, which are transformed into embeddings by the embedding layer. The embedding layer, typically a matrix of vocabulary size by hidden dimension, functions as a lookup table that retrieves an embedding for each index. In conventional settings, the parameters of embedding layer are learnable and updated during training. }
  \label{fig:ehrfm}
\end{figure}

The target outcomes for finetuning included three categories: clinical outcomes \citep{guo2023ehr}, patient phenotypes \citep{harutyunyan2019multitask}, and in-hospital events. Clinical outcomes included in-hospital mortality (MT), long length of stay (LLOS), and readmission (RA). Patient phenotypes (Pheno) included 24 acute care conditions. In-hospital events consisted of urinary tract infection (UTI) \citep{moller2021prediction}, fracture (Fx) \citep{almog2020deep}, sepsis \citep{nemati2018interpretable}, pneumonia (PNA) \citep{effah2022machine}, and myocardial infarction (MI) \citep{mandair2020prediction}. The incidence rates of these target outcomes are summarized in Table \ref{tab:characteristics}. Among the clinical outcomes, LLOS had the highest incidence rates in both hospitals, while MT and RA rates were higher in EHRSHOT than in MIMIC-IV. For phenotypes, the incidence rates of all phenotypes were higher than 2\% for both datasets. Incidence rates of all in-hospital events were higher in EHRSHOT than MIMIC-IV. All target outcomes were labeled based on definitions from previous studies \citep{guo2023ehr, harutyunyan2019multitask, kim2025pretrained}. Prediction timepoints were defined as follows: for MT and LLOS, predictions were made at midnight on the day of admission; for RA, at midnight on the day of discharge; for phenotypes, at midnight on the second day of admission; and for in-hospital events, one day prior to their occurrence.

\begin{figure*}[!t]
  \centerline{\includegraphics[width=\textwidth]{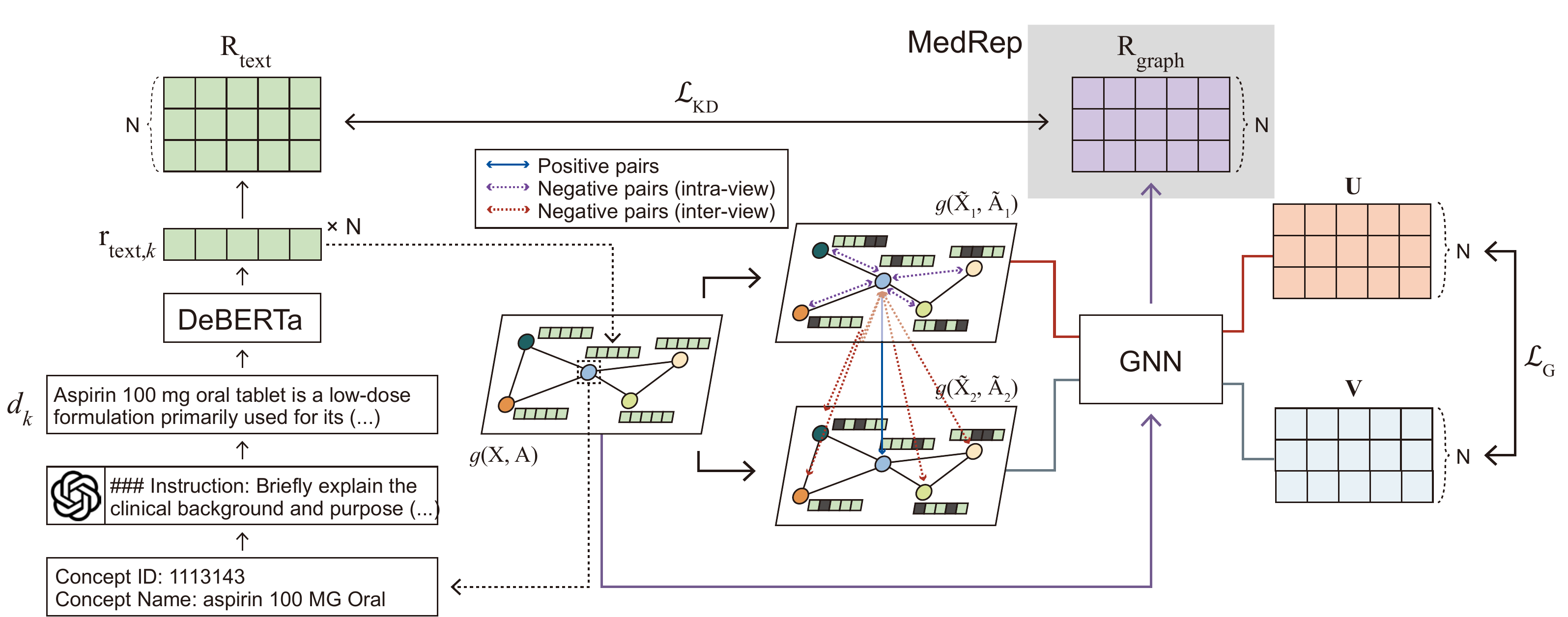}}
  \caption{Medical concept representation learning process. For $k$-th medical concept with minimal textual definition, LLM-attributed description $d_k$ is assigned. The description $d_k$ is then encoded into a representation $r_{\text{text},k}$ by a small language model (e.g., DeBERTa). Each $r_{\text{text},k}$ serves a node feature of OMOP vocabulary relational graph. MedRep is trained by iteratively updating $r_{\text{text},k}$ using both graph contrastive learning loss $\mathcal{L}_\mathrm{G}$ and knowledge distillation loss $\mathcal{L}_{\mathrm{KD}}$. $\mathcal{L}_{\mathrm{KD}}$ is applied to prevent MedRep from forgetting information acquired from LLM-attributed descriptions.}
  \label{fig:medrep}
\end{figure*}

\paragraph{Medical concept representation (MedRep)} MedRep is a set of medical concept representations. An EHR foundation model typically consists of embedding layers, self-attention layers, and subsequent components for pretraining or finetuning (Figure \ref{fig:ehrfm}). The input to EHR foundation models is a sequence of medical code indices, which is called a patient trajectory, along with corresponding additional information such as age and time. Separate embedding layers exist for medical codes and for each additional information. Each embedding layer is a lookup table that maps an index to a numerical vector (embedding). For each point, all embeddings (e.g., medical code, age, and time) are summed and then pass through self-attention layers and subsequent pretraining or finetuning layers according to the objective. During pretraining or finetuning, all parameters of EHR foundation models, including the embedding layers, are updated. Each embedding serves as a representation of the corresponding index. In contrast, our approach replaces the learnable embedding layer for medical codes with fixed, pretrained representations, MedRep.  

Using phrase-level concept names and the graph ontology of OMOP vocabulary, we first created text-based representations by leveraging a language model along with LLM-attributed descriptions for each concept. These text-based representations were then enhanced by relational information between concepts, utilizing the graph ontology of OMOP vocabulary and graph contrastive learning (GCL) \citep{zhu2020deep}. For GCL, we used the GRACE framework, where two different graph views are generated to optimize the pairwise objective, arranging positive pairs to be close and negative pairs to be apart \citep{zhu2020deep}.

\begin{figure*}[!htb]
  \centerline{\includegraphics[width=\textwidth]{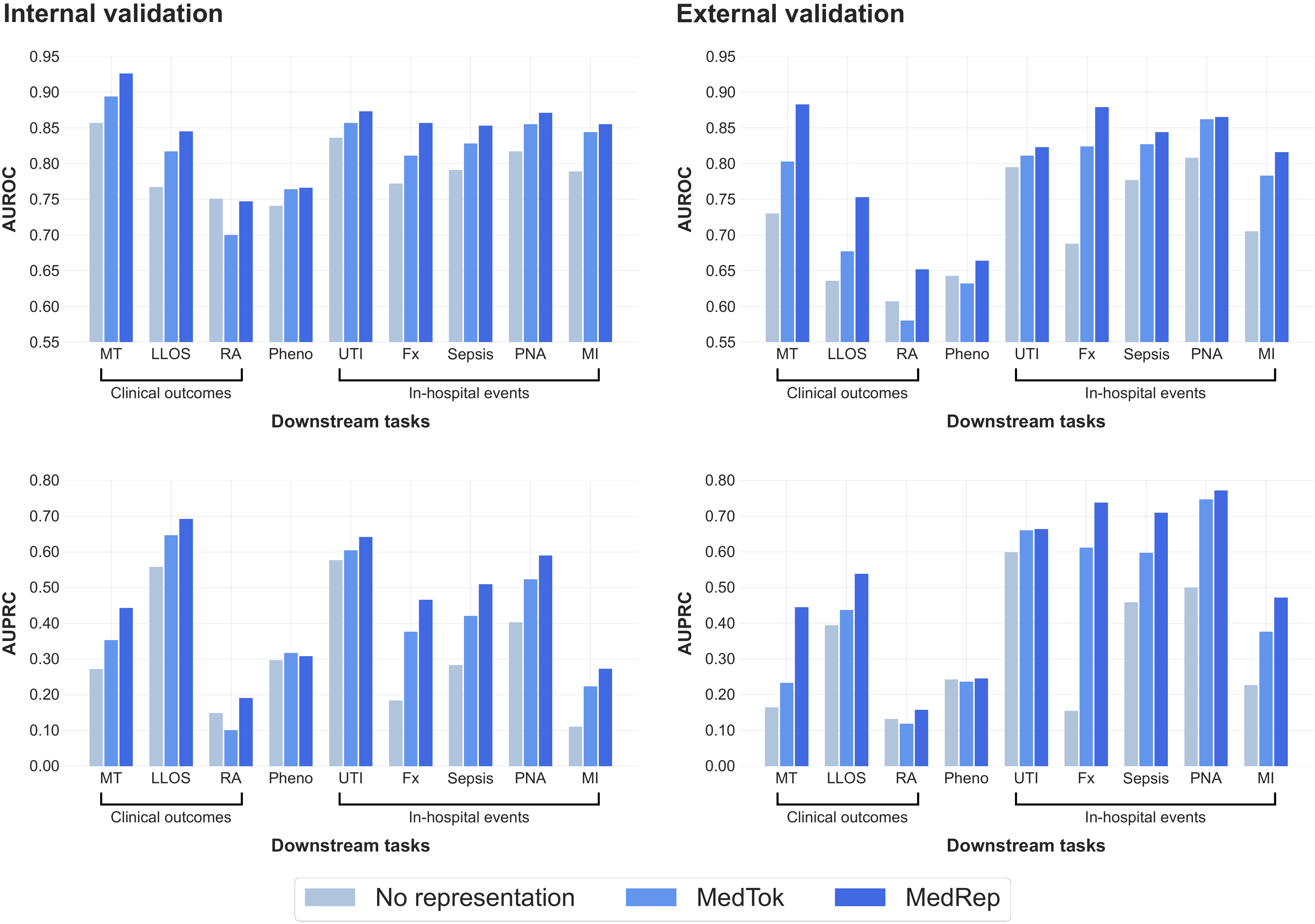}}
  \caption{Downstream task performance. All downstream tasks were performed with three random seeds, and the reported AUROC and AUPRC values were averaged across these runs. Phenotype prediction is multilabel classification problem; therefore, we reported macro AUROC and macro AUPRC. All other tasks are binary classification problems.}
  \label{fig:performance}
\end{figure*}

However, we observed that simply applying GCL led to representations that are extremely sparse and less informative. This phenomenon was because text-based representations were easily separable using only a few features, and our graph neural network (GNN) in GCL focused primarily on distinguishing positive and negative pairs rather than enriching current representations. This resulted in oversimplified embeddings that are easily separable but lack semantic detail. To address this problem, we adopted a knowledge distillation (KD) loss from Learning without Forgetting (LwF) \citep{li2017learning}, a continual learning method designed to prevent catastrophic forgetting \citep{goodfellow2015empiricalinvestigationcatastrophicforgetting, MCCLOSKEY1989109}. MedRep was trained by iteratively minimizing the GCL loss and the KD loss, thereby capturing relational information from the OMOP graph ontology while preserving previously learned textual information. A brief illustration of our medical concept representation learning process is shown in Figure \ref{fig:medrep}. 

Since MedRep is fixed and contains the clinical information of each concept, it not only improves the performance of EHR foundation models but also offers flexibility in coping with unseen concepts. We included 7,572,911 medical concepts, covering all medical codes from the condition, drug, measurement, and procedure domains of OMOP CDM. These concepts span 66 medical vocabularies, including SNOMED-CT \citep{donnelly2006snomed}, ICD-10 \citep{quan2005coding}, ICD-9 \citep{quan2005coding}, RxNorm \citep{liu2005rxnorm}, NDC \citep{united1969national}, LOINC \citep{mcdonald2003loinc}, and HCPCS \citep{nusgart2018tips}.

\paragraph{Downstream task performance} We evaluated the downstream task performance of EHR foundation models across three settings: without pretrained representations (vanilla), with MedTok, and with MedRep (Figure \ref{fig:performance}). BEHRT \citep{li2020behrt} was used as the baseline architecture for the EHR foundation model. MedTok generally outperformed vanilla models in both internal and external validations, except for all RA and phenotype prediction tasks in external validation. MedRep achieved the best overall performance across all downstream tasks, except for the area under the receiver operating characteristic curve (AUROC) for RA and the area under the precision-recall curve (AUPRC) for phenotype prediction in internal validation. Detailed performance metrics for all downstream tasks are summarized in Supplementary Table 1.

The performance improvements achieved by MedTok and MedRep are summarized in Table \ref{tab:perfimp}. Both methods enhanced the performance of EHR foundation models across most downstream tasks, with improvements mostly greater in external validation than in internal validation. This demonstrates that pretrained representations enhance not only the model performance but also its adaptability to new data. The improvements with MedRep were consistently greater than those with MedTok, indicating that MedRep can be a more effective option to enhance the generalizability of EHR foundation models.

\begin{table*}[!]
\centering
  \caption{Performance improvements achieved by MedTok and MedRep. Bold indicates the higher improvements.}
\begin{tabular}{llllllllllll}
\toprule
 & &  \textbf{MT} & \textbf{LLOS} & \textbf{RA} & \textbf{Pheno} & \textbf{UTI} & \textbf{Fx} & \textbf{Sepsis} & \textbf{PNA} & \textbf{MI} & \textbf{Avg} \\
 \midrule
\multicolumn{2}{l}{\textit{Metric: AUROC}} &  &  &  &  &  &  &  &  &  &  \\
Model & Validation &  &  &  &  &  &  &  &  &  &  \\
 \midrule
MedTok & Internal & 0.037 & \textbf{0.05} & -0.051 & \textbf{0.023} & \textbf{0.021} & 0.039 & 0.037 & 0.038 & 0.055 & 0.028 \\
 & External & \textbf{0.073} & 0.041 & \textbf{-0.027} & -0.011 & 0.016 & \textbf{0.136} & \textbf{0.05} & \textbf{0.054} & \textbf{0.078} & \textbf{0.046} \\
 \midrule
MedRep & Internal & 0.069 & 0.078 & -0.004 & \textbf{0.025} & \textbf{0.037} & 0.085 & 0.062 & 0.054 & 0.066 & 0.052 \\
 & External & \textbf{0.153} & \textbf{0.117} & \textbf{0.045} & 0.021 & 0.028 & \textbf{0.191} & \textbf{0.067} & \textbf{0.057} & \textbf{0.111} & \textbf{0.088} \\
 \midrule
\multicolumn{2}{l}{\textit{Metric: AUPRC}} &  &  &  &  &  &  &  &  &  &  \\
Model & Validation &  &  &  &  &  &  &  &  &  &  \\
 \midrule
MedTok & Internal & \textbf{0.081} & \textbf{0.089} & -0.048 & \textbf{0.02} & 0.028 & 0.193 & 0.138 & 0.121 & 0.113 & 0.082 \\
 & External & 0.068 & 0.043 & \textbf{-0.013} & -0.006 & \textbf{0.062} & \textbf{0.457} & \textbf{0.139} & \textbf{0.246} & \textbf{0.15} & \textbf{0.127} \\
 \midrule
MedRep & Internal & 0.171 & 0.135 & \textbf{0.042} & \textbf{0.011} & 0.065 & 0.282 & 0.227 & 0.187 & 0.162 & 0.142 \\
 & External & \textbf{0.28} & \textbf{0.144} & 0.026 & 0.003 & 0.065 & \textbf{0.583} & \textbf{0.251} & \textbf{0.271} & \textbf{0.245} & \textbf{0.208} \\
 \bottomrule
\end{tabular}
  \label{tab:perfimp}
\end{table*}

\begin{figure}[!htb]
  \centerline{\includegraphics[width=\columnwidth]{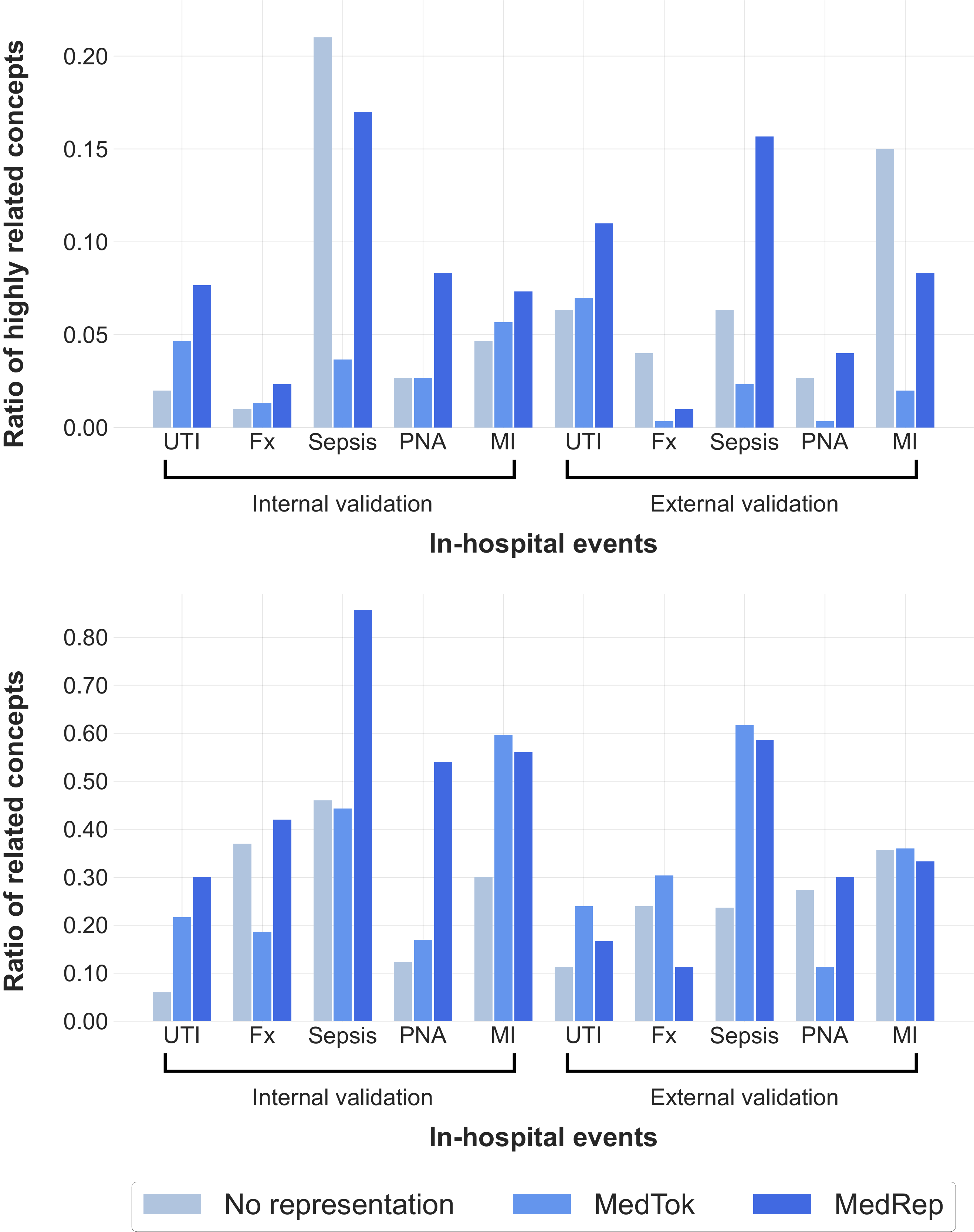}}
  \caption{Ratio of highly related and related concepts. An LLM categorized the relevance of given concepts to the target outcome as \textit{low}, \textit{moderate}, and \textit{high}. Highly related concepts include those labeled with \textit{high}, and related concepts include those labeled with \textit{high} or \textit{moderate}. As we conducted experiments with three random seeds, we reported the average ratio.}
  \label{fig:qual}
\end{figure}

\begin{table*}[!]
\centering
  \caption{Performance variation with graph ontology. Bold indicates the best performance.}
\begin{tabular}{llllllllllll}
\toprule
 & &  \textbf{MT} & \textbf{LLOS} & \textbf{RA} & \textbf{Pheno} & \textbf{UTI} & \textbf{Fx} & \textbf{Sepsis} & \textbf{PNA} & \textbf{MI} & \textbf{Avg} \\
 \midrule
\multicolumn{2}{l}{\textit{Internal validation}} &  &  &  &  &  &  &  &  &  &  \\
Metric & Model &  &  &  &  &  &  &  &  &  &  \\
 \midrule
AUROC & Vanilla & 0.857 & 0.767 & \textbf{0.751} & 0.741 & 0.836 & 0.772 & 0.791 & 0.817 & 0.789 & 0.791 \\
 & No graph & \textbf{0.926} & \textbf{0.845} & 0.737 & \textbf{0.771} & 0.86 & 0.848 & \textbf{0.861} & 0.869 & \textbf{0.871} & 0.843 \\
 & No LwF & 0.916 & 0.823 & 0.727 & 0.765 & 0.86 & 0.85 & 0.846 & 0.856 & 0.84 & 0.831 \\
 & MedRep & \textbf{0.926} & \textbf{0.845} & 0.747 & 0.766 & \textbf{0.873} & \textbf{0.857} & 0.853 & \textbf{0.871} & 0.855 & \textbf{0.844} \\
 \midrule
AUPRC & Vanilla & 0.272 & 0.558 & 0.149 & 0.297 & 0.577 & 0.184 & 0.283 & 0.403 & 0.111 & 0.315 \\
 & No graph & 0.439 & 0.685 & 0.099 & \textbf{0.32} & 0.622 & 0.434 & \textbf{0.519} & 0.573 & \textbf{0.281} & 0.441 \\
 & No LwF & 0.364 & 0.659 & 0.137 & 0.303 & 0.628 & 0.439 & 0.491 & 0.565 & 0.213 & 0.422 \\
 & MedRep & \textbf{0.443} & \textbf{0.693} & \textbf{0.191} & 0.308 & \textbf{0.642} & \textbf{0.466} & 0.51 & \textbf{0.59} & 0.273 & \textbf{0.457} \\
 \midrule
\multicolumn{2}{l}{\textit{External validation}} &  &  &  &  &  &  &  &  &  &  \\
Metric & Model &  &  &  &  &  &  &  &  &  &  \\
 \midrule
AUROC & Vanilla & 0.73 & 0.636 & 0.607 & 0.643 & 0.795 & 0.688 & 0.777 & 0.808 & 0.705 & 0.71 \\
 & No graph & 0.864 & 0.735 & 0.621 & 0.66 & 0.801 & 0.86 & \textbf{0.858} & \textbf{0.869} & \textbf{0.826} & 0.788 \\
 & No LwF & 0.778 & 0.7 & 0.596 & 0.635 & 0.797 & 0.864 & 0.8 & 0.855 & 0.75 & 0.753 \\
 & MedRep & \textbf{0.883} & \textbf{0.753} & \textbf{0.652} & \textbf{0.664} & \textbf{0.823} & \textbf{0.879} & 0.844 & 0.865 & 0.816 & \textbf{0.798} \\
 \midrule
AUPRC & Vanilla & 0.165 & 0.395 & 0.132 & 0.243 & 0.599 & 0.155 & 0.459 & 0.501 & 0.227 & 0.32 \\
 & No graph & 0.383 & 0.512 & 0.136 & \textbf{0.256} & 0.647 & 0.675 & \textbf{0.729} & 0.745 & \textbf{0.505} & 0.51 \\
 & No LwF & 0.273 & 0.474 & 0.12 & 0.228 & 0.645 & 0.664 & 0.612 & 0.76 & 0.351 & 0.459 \\
 & MedRep & \textbf{0.445} & \textbf{0.539} & \textbf{0.158} & 0.246 & \textbf{0.664} & \textbf{0.738} & 0.71 & \textbf{0.772} & 0.472 & \textbf{0.527} \\
 \bottomrule
\end{tabular}
  \label{tab:graph}
\end{table*}

\paragraph{Impact of graph ontology} For MedRep, relational information among medical concepts was additionally learned by integrating the graph ontology of OMOP vocabulary with LLM-attributed text-based representations. As shown in Table \ref{tab:graph}, MedRep without the graph ontology (i.e., solely text-based representations) still outperformed vanilla models in both internal and external validation. Overall performance declined when MedRep was trained without LwF; however, MedRep trained with GCL combined with LwF improved average performance, especially in external validation. Specifically, MedRep exhibited consistent improvements in predicting clinical outcomes (MT, LLOS, RA), UTI, and Fx, while sepsis and MI were better with MedRep without the graph ontology.

\paragraph{Qualitative analysis} We conducted a qualitative analysis using finetuned models for in-hospital events, as these outcomes generally have identifiable causes under common clinical backgrounds, whereas MT, LLOS, RA, and phenotypes are influenced by a wide variety of factors. Specifically, we extracted the top 100 most important concepts based on the attention scores of finetuned models, and used MedGemma-27B \citep{sellergren2025medgemmatechnicalreport} to classify their relevance to the target outcomes as low, moderate, or high. As shown in Figure \ref{fig:qual}, models with MedRep leveraged a greater number of highly related concepts (labeled as high) for prediction compared to the other baselines, in six out of nine outcomes. Both MedTok- and MedRep-based models used a greater number of the related concepts (labeled as high or moderate) than the vanilla models.

\begin{figure}[!htb]
  \centerline{\includegraphics[width=\columnwidth]{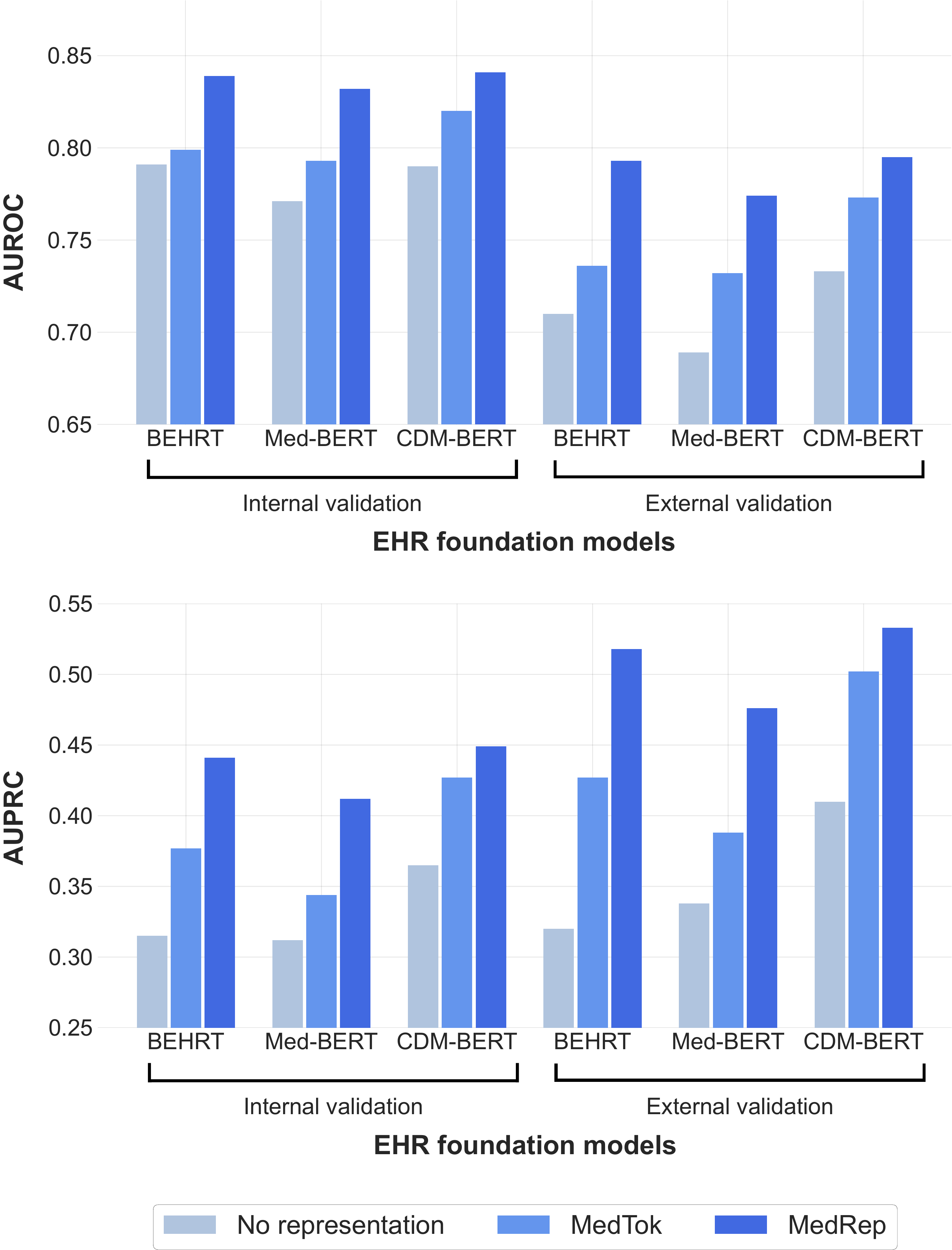}}
  \caption{Downstream task performance on other EHR foundation models. All downstream tasks were performed with three random seeds, and the reported AUROC and AUPRC values were averaged across all experiments.}
  \label{fig:scale}
\end{figure}

\paragraph{Scalability of MedRep} To demonstrate the effectiveness of MedRep across diverse EHR foundation models, we further conducted experiments with Med-BERT \citep{rasmy2021med} and CDM-BERT \citep{kim2025pretrained}. Med-BERT additionally adopts a prolonged length of stay prediction process during pretraining to enhance the generalizability of the pretrained model. CDM-BERT introduces domain embedding to indicate the domain of masked tokens during pretraining, to prevent the model from finding unnecessary medical concepts from other domains. For both Med-BERT and CDM-BERT, MedRep achieved the best overall performance, followed by MedTok (Figure \ref{fig:scale}). MedRep is applicable to any EHR foundation model that utilizes medical codes. Detailed performance values are summarized in Supplementary Tables 2 and 3.

\section{Discussion}
In this study, we introduced MedRep, a set of OMOP medical concept representations that can be directly integrated into any EHR foundation model to improve performance and generalizability. By leveraging an LLM and the graph ontology of OMOP vocabulary, we generated comprehensive medical concept representations. To address the oversimplification issue caused by GCL, we adopted a continual learning approach by incorporating the KD loss from LwF to retain knowledge learned from the LLM-attributed descriptions. MedRep consistently improved performance across diverse clinical tasks, especially in external validation. Furthermore, MedRep effectively identified features clinically related to target outcomes, enhancing the explainability of EHR foundation models. Finally, we demonstrated consistent performance of MedRep across diverse EHR foundation models.

Nowadays, most medical institutions are adopting EHR systems, and enormous amounts of healthcare data are accumulated every day \citep{liang2021adoption}. Despite this abundance, EHR data is generally not accessible to external researchers, and exporting EHR data is hardly available due to patient privacy regulations \citep{aliberti2022}. As a result, even though there have been a lot of studies that utilized EHR foundation models, they have typically utilized data from no more than three institutions \citep{guo2024multi}. National claim data covering broader populations have been used in some cases; however, they lack detailed clinical observations such as continuous measurements \citep{gliklich2019tools}. To summarize, hospital EHR data is dense and high quality but small in scale and hardly sharable, whereas national claim data is large in scale but relatively low in quality. To enhance the generalizability and quality of EHR-based machine learning models, it is essential to merge the knowledge from multiple hospital datasets. In this context, MedRep offers an optimal solution, as it encourages similar concepts to have similar representations even if they have not been observed simultaneously. In addition, OMOP CDM has been applied to a lot of medical institutions \citep{reinecke2021usage, biedermann2021standardizing}, and machine learning models trained by OMOP CDM-based EHR data can be shared across institutions, because exporting only the weights of the model parameters is typically not restricted. Through MedRep, EHR foundation models developed at different institutions can cooperate to achieve general performance for various medical tasks. Therefore, MedRep can serve as a standard baseline representation for a large EHR foundation model. 

As a prior work, MedTok \citep{su2025multimodal} has been proposed to quantize original medical codes into a unified codebook. MedTok is based on textual information for more than 600,000 medical codes from eight vocabularies and on relational knowledge from PrimeKG \citep{chandak2023building}. MedTok first generates text semantic and graph-level embeddings for modality-specific and cross-modality embeddings. Then, MedTok learns the codebook using those two embedding types and provides unified medical tokens for EHR foundation models. On the contrary, MedRep learns representations of medical concepts using LLM-attributed textual information and OMOP graph ontology. MedRep consists of more than 7.5 million medical concepts from 66 medical vocabularies, which can be mapped from/to OMOP concepts, including measurement concepts such as laboratory tests and vital signs, which are not included in MedTok. The influence of graph ontology in MedRep was weaker than in MedTok; while MedTok aligns the representations from text and graph to be similar, we first trained text-based representations and then refined them using GCL and KD losses. This was because the relational information from OMOP vocabulary is limited to mapping and hierarchical information rather than the clinical relationship between concepts, whereas PrimeKG contains richer clinical relational information but does not cover all OMOP concepts. Our experiments exhibit that MedRep outperforms MedTok, demonstrating the effectiveness of MedRep in the OMOP CDM environment. 

Although the influence of relational information was weaker than in the previous work \citep{su2025multimodal}, the overall performance was greater with graph ontology, especially for predicting clinical outcomes. This suggests that medical concept representations enhanced with relational information can better capture factors contributing to complex outcomes. Meanwhile, the predictive performance of MedRep without the graph ontology was still comparable, allowing users to choose between the full MedRep and the graph-free version depending on their needs. Additionally, we experimented with combining the representations from MedRep and MedRep without a graph by summation, but this did not improve the performance.

One of the current challenges in EHR foundation models is explainability. Since EHR foundation models are based on self-attention architectures, most studies have provided qualitative analysis using attention scores from trained models, often supplemented by insights from clinicians or prior clinical research \citep{huang2023development, li2020behrt, kim2025pretrained}. However, seeking advice from clinicians is costly and hardly available in real time, because clinicians should review the enormous records of patients. To address this problem, we used MedGemma-27B, a state-of-the-art LLM designed to interpret and reason about medical images and text \citep{sellergren2025medgemmatechnicalreport}. For instance, MedGemma-27B identified the following concept IDs as highly related to UTI: 3004501 (Glucose [Mass/volume] in Serum or Plasma), 3037244 (Yeast [\#/area] in Urine sediment by Microscopy high power field), 3004410 (Hemoglobin A1c/Hemoglobin.total in Blood), 4328657 (Color of Urine), 3022621 (pH of Urine by Test strip), 197320 (Acute kidney injury), 4193704 (Type 2 diabetes mellitus without complication), and 3021601 (Nitrite [Presence] in Urine by Test strip). All these concepts are well-established in the literature as closely associated with UTI \citep{schersten1967subnormal, guze1958fungus, patterson1997bacterial, levey2017acute}. While LLMs are excellent tools for providing general and comprehensible clinical information for users, they often struggle to reflect abundant clinical information from patient-specific longitudinal EHR data. On the other hand, EHR foundation models are adept at capturing individual complex medical histories for diverse specific clinical tasks, but still face challenges in explainability. Considering MedGemma effectively identifies relevant concepts from EHR foundation models, MedRep-based EHR foundation models with LLM may create strong synergy, complementing each other. 

This study has several limitations. First, we utilized only publicly available datasets, MIMIC-IV and EHRSHOT. It was to encourage reproducibility, but for further demonstration, it is essential to implement our approach to larger real-world datasets encompassing more diverse patient populations. Second, we did not use the observation table from OMOP CDM. The observation table contains clinical facts obtained in the context of examination, questioning, or a procedure, and any data that cannot be represented by existing domains, such as social and lifestyle facts, medical history, and family history. The data of the observation table usually consists of unstructured plain texts, and there is no standard for mapping those various records into standardized concepts. Representing and integrating these records into EHR foundation models remains future work. Third, we employed only the GRACE framework for GCL while there exist various other GCL methods. While the improvement by the graph ontology was observed, the gains were not significant across all tasks, indicating the need to investigate alternative approaches to better reflect relational information. Fourth, the downstream tasks were traditional, and the cohort definitions were not strictly defined by clinical experts. Further validation is required to determine if MedRep can be effective in more complex clinical tasks with strictly defined cohorts. 

We suggest future work as follows: First, since MedRep covers dozens of medical vocabularies, developing dictionaries or tools that retrieve the appropriate representation for each code would be valuable. Second, exploring strategies to merge EHR foundation models from multiple institutions is important, as MedRep can standardize code systems across models, and many institutions are adopting OMOP CDM. Scaling laws \citep{kaplan2020scaling} for EHR foundation models can be explored through the unified code system. Third, integrating PrimeKG with OMOP CDM could help enrich currently limited relational information of MedRep, as PrimeKG contains more clinically detailed relationships. Fourth, clinical context and hierarchical information contained in MedRep could be utilized beyond EHR foundation models, such as enhancing claim code systems used in hospitals and insurance, supporting healthcare policy development. Fifth, report-generating models for unstructured medical data \citep{liu2019clinically, wan2024meit} can be integrated into MedRep. Instead of simply concatenating the representations from multiple modalities, utilizing generated reports to create representations aligned with MedRep can contribute to building a multimodal EHR foundation model.

\section{Methods}
\paragraph{Concept curation}
For OMOP concepts, we extracted concept IDs from the central OHDSI vocabularies repository, Athena (\url{https://athena.ohdsi.org/search-terms/start}), which is regularly updated by the OHDSI community  \citep{reinecke2021usage}. We selected 7,572,911 concepts, each with a concept name, from the drug, condition, measurement, and procedure domains, covering 66 medical vocabularies (summarized in Supplementary Table 4). Concepts from the condition, drug, and procedure were tokenized into their corresponding indices. For measurement concepts without numerical value, they were tokenized in the same way as for condition, drug, and procedure concepts. Each measurement concept with numerical values was divided into deciles, and the corresponding decile number (0-9) was appended to both the concept ID and the concept name. Finally, the dataset was expanded to 7,730,741 concepts after adding measurement concepts with deciles. 

For the OMOP graph ontology, we excluded concepts with more than 100 links, as such concepts tend to be overly broad. These codes can be easily included via neighbor sampling, which may unintentionally allow marginal message passing during GCL. Examples of excluded nodes include dose form codes such as 46234469 (Injection) and 19082168 (Oral Capsule), as well as procedure group codes such as 42793831 (Surgery acts) and 42793832 (Act of anesthesia). 

The final graph consisted of 5,129,561 nodes and 9,675,551 edges. Each edge has properties such as ``is a'', ``subsumes'', and ``mapped from'', but for simplicity in implementing GCL, we regarded all edges as unattributed. 

\paragraph{Prompt design for generating descriptions} We enriched the clinical information of the given concepts using an LLM \citep{schick2020exploiting}, because each OMOP concept has only minimal definitions, which are insufficient for generating representation with language models. In addition, the OMOP graph ontology generally includes hierarchical and mapping information between concepts, rather than cross-domain clinical relationships, such as which drugs are associated with specific conditions. Thus, the objective of generating LLM-attributed descriptions was to supplement each concept with richer textual clinical information, focusing on its cause, purpose, or context. For all domains, we instructed the LLM to describe the clinical background of each concept. Specifically, we requested information on treatments for conditions; the purpose and clinical meaning of ingredient, dosage form, and strength for drugs; the interpretation of each decile for measurements; and the clinical purpose of procedures. For drug concepts containing multiple substances, such as 45774969 (netupitant 300 MG / palonosetron 0.5 MG Oral Capsul), we requested comprehensive explanations covering all included drugs. We used ChatGPT-4o-mini (OpenAI Inc) for the LLM. The prompt for each domain and sample descriptions are provided in Supplementary Table 5. 

\paragraph{MedRep: Medical concept representation learning} For text-based representations, we trained a small language model based on masked language model (MLM) \citep{devlin2019bert} objective using the generated descriptions to encapsulate enriched clinical information. For each concept, the final text-based representation is denoted as $\mathbf{R}_{\text{text}}\in\mathbb{R}^{N\times h}$ where $N$ is the vocabulary size and $h$ is hidden dimension of the representation. We used DeBERTa \citep{he2020deberta} for small language model and enlarged the context length from 512 to 2048 tokens. The model was trained with a batch size of 32 and a learning rate of 0.00005. We split $\mathbf{R}_{\text{text}}$ into training and validation sets at an 8:2 ratio and validated the MLM loss every 200 training batches using 100 validation batches. Pretraining was terminated by early stopping with the patience of 30. 

Even though similar concepts are expected to have similar description and similar text-based representations, we further refined these representations using the officially verified graph ontology of the OMOP vocabulary. Let the relational graph be $G=\left(C,E\right)$, where $C=\left\{c_k\right\}_{k=1}^N$ denotes the set of concepts (nodes) and $E\ \subseteq C\ \times C$ indicates the set of edges. The feature matrix and adjacency matrix are denoted as $X\in\mathbb{R}^{N\times h}$ and $A\in\left\{0,\ 1\right\}^{N\times N}$, where $X_k=r_{text,k}$ is the text-based representation of the $k$-th concept, and $A_{ij}=1$ if and only if $\left(c_i,c_j\right)\in E$. Our objective at this step was to train a GNN encoder $g$ to produce final concept representations $R_{graph}=g\left(\mathbf{X},\ \mathbf{A}\right)\in\mathbb{R}^{N\times h}$, as illustrated in Figure 2.

For GCL, we adopted the GRACE framework \citep{zhu2020deep}, which is a general contrastive learning framework for unsupervised graph representation learning. GRACE generates two different graph views to obtain two embedding sets $U=g(\widetilde{\mathbf{X}_1},\widetilde{\mathbf{A}_1})$ and $V=g(\widetilde{\mathbf{X}_2},\widetilde{\mathbf{A}_2})$, where $\widetilde{\mathbf{X}}$ and $\widetilde{\mathbf{A}}$ are masked node features and dropped edges. We define the critic as $\theta\left(u,v\right)=s\left(p\left(u\right),p\left(v\right)\right)/\tau$, where $s$ is the cosine similarity, $p$ is a non-linear projection, and $\tau$ is a temperature parameter. The pairwise objective arranging positive pairs to be close and negative pairs to be apart for each positive pair $\mathbf{u}_k\in U$ and $\mathbf{v}_k\in V$ is
$$\ell(\mathbf{u}_k, \mathbf{v}_k) = \log \frac{e^{\theta(\mathbf{u}_k, \mathbf{v}_k)}}{
\underbrace{e^{\theta(\mathbf{u}_k, \mathbf{v}_k)}}_{\text{positive pair}}
+ \sum\limits_{i \ne k}{\underbrace{\left( e^{\theta(\mathbf{u}_k, \mathbf{v}_i)} + e^{\theta(\mathbf{u}_k, \mathbf{u}_i)} \right)}_{\text{inter- and intra-view negative pairs}}
}},$$
and the parameters of GNN encoder $g$ are updated with the following objective to be minimized: 
$$\mathcal{L}_\text{G}=-\frac{1}{2N}\sum_{k=1}^{N}{\left[\ell\left(\mathbf{u}_k,\mathbf{v}_k\right)+\ell\left(\mathbf{v}_k,\mathbf{u}_k\right)\right].}$$
The configuration for GRACE in this study is summarized in Supplementary Table 6. As representation learning involved more than 7.7 million concepts, allocating the whole feature matrix and adjacency matrix in memory was not possible. Instead, for each batch, we constructed a subgraph by extracting 30, 20, and 10 neighbors for 1-hop, 2-hop, and 3-hop connections from each node in the batch. 

While GCL enhances the text-based representations via the OMOP graph ontology, the text-based representations are already highly informative, and the GNN might easily distinguish concepts resulting in loss of information \citep{zhang2023mitigating}. To mitigate this problem, we additionally trained the GNN with the scheme of LwF, a continual learning approach for maintaining previously learned information when training with new data. As a knowledge distillation loss $\mathcal{L}_{\mathrm{KD}}$, we adopted Kullback–Leibler (KL) divergence loss between the original text-based representations $\mathbf{R}_{\text{text}}$ and the GNN-attributed representations $\mathbf{R}_{\text{graph}}$ as follows: $$\mathcal{L}_{\mathrm{KD}}(\mathbf{R}_{\mathrm{text}}\parallel \mathbf{R}_{\mathrm{graph}})=\sum_{k=1}^{N}S\left(r_{\mathrm{text},k}\right)\cdot\log{\left(\frac{S\left(r_{\mathrm{text},k}\right)}{S\left(r_{\mathrm{graph},k}\right)}\right)},$$
where $\mathbf{R}_{\mathrm{text}}=\left\{r_{\mathrm{text},k}\right\}_{k=1}^N$, $\mathbf{R}_{\mathrm{graph}}=\left\{r_{\mathrm{graph},k}\right\}_{k=1}^N=g\left(\mathbf{R}_{\mathrm{text}},\mathbf{A}\right)$
 and $S$ is the softmax function. The parameters of GNN encoder $g$ were alternately updated by minimizing $\mathcal{L}_G\ and \mathcal{L}_{\mathrm{KD}}$. As a result, we obtained the final concept representation set $\mathbf{R}_{\mathrm{graph}}$. EHR foundation models can directly use pretrained representations instead of training the embedding layer for medical concept indices.

\paragraph{Trajectory construction} Each patient trajectory contains integer sequences of medical concept indices, age indices, visit indices, record indices, and domain indices, which are sorted in order of time. Each medical concept and age index refer to the corresponding concept and age. Each visit is assigned a unique visit index starting from 1 for each patient, and within each visit, the record index increments by 1 for each day, also starting from 1. The record index serves to encode temporal information for long visits. For CDM-BERT, domain index (ranging from 0 to 4) is assigned to every concept according to its domain (special token, condition, drug, measurement, and procedure). The embeddings of the concept itself, age, visit, record, and domain (for CDM-BERT) are summed and fed into the following layers. The maximum trajectory length was set to 2048. Every trajectory starts with the [CLS] token, and the [SEP] token was not used. Trajectories were padded by the [PAD] token with the maximum length. 

\paragraph{Pretraining and finetuning details} We used three EHR foundation models, BEHRT \citep{li2020behrt}, Med-BERT \citep{rasmy2021med}, and CDM-BERT \citep{kim2025pretrained}. For pretraining, the trajectories with more than 2048 concepts are sliced into non-overlapping sub-trajectories with the maximum length to prevent potential dependency among trajectories from a single patient. The batch size and learning rate were set to 32 and 0.00005, respectively. We selected the pretraining models with the lowest MLM loss for 50 epochs with masked ratio of 0.3. AdamW optimizer \citep{loshchilov2017decoupled} was adopted for parameter update with a weight decay of 0.01. Pretraining was executed with 8 NVIDIA RTX A6000 GPUs for 2-3 days for each model. 

For finetuning, the records before the prediction timepoint were utilized for prediction tasks, with the maximum length of trajectory 2048. For pretrained baseline models, a fully connected layer-based classification head was combined to predict the outcomes. We validated the model performance for every epoch and if the performance did not increase for 5 times, the finetuning process was terminated with early stopping. Each finetuning task was executed with a single NVIDIA RTX A6000 GPU for 4-8 hours. All finetuning tasks were performed with three random seeds, randomly splitting the training, validation, test sets of MIMIC-IV. 

\paragraph{Outcome labeling}
For MT and LLOS, we used the last visit for each patient, excluding those who died on the day of admission. LLOS was labeled for patients hospitalized more than one week. For RA, we randomly selected a visit except for the last visit and labeled patients who were hospitalized within one month after discharge. 
For phenotypes, since previous studies labeled them based on ICD-9 codes, we converted the code set for each phenotype into OMOP concepts, including all child codes. The prediction timepoint was set to midnight on the second day of admission. We selected the last visit with at least two days of hospitalization and excluded all patients who had at least one phenotype before the prediction timepoint. 
For each in-hospital event, including UTI, Fx, Sepsis, PNA, and MI, we first sorted the patients who had never suffered from the target outcome to form the control group and randomly selected a prediction timepoint for them. For the case group, we also randomly selected the day before the target outcome. For RA and in-hospital events, prediction timepoints were randomly assigned to minimize bias caused by our criteria.

\paragraph{MedTok training}
MedTok requires medical code description and local subgraph for each medical code. However, since our study used a different EHR format and medical code set, we generated MedTok using our pretrained text-based representations and the relational graph from the OMOP vocabulary, instead of using description text and PrimeKG. In our setting, covering OMOP concepts using PrimeKG, which contains around 130,000 nodes, was not feasible. 

\paragraph{Qualitative analysis}
For qualitative analysis, we extracted the top 100 most important concepts from each in-hospital event prediction model. Specifically, for each patient, we obtained attention scores from the [CLS] token of the last self-attention layer. We then identified the top 100 concepts with the highest attention scores and counted how frequently each concept appeared in these top 100 lists across patients. Finally, we selected the 100 most frequently occurring important concepts for each model. Since all downstream tasks were run with three different random seeds, we combined the results and obtained the top 300 concepts for each in-hospital event. 
To evaluate the clinical relevance of these concepts, we instructed an LLM to classify each concept’s relevance to the target outcomes as low, moderate, or high. The prompt used was: “We trained a machine learning model to predict [outcome] based on patients' past medical records. From this model, we extracted 100 important features. Now, we want to assess how strongly each of these features is related to [outcome]. Please evaluate each feature and categorize its relevance to [outcome] as one of the following: "Low", "Moderate", or "High". Make sure to provide an assessment for all 100 features — do not omit any. Please provide your answer in CSV format with two columns: feature, assessment. The extracted features are as follows: [100 extracted concepts]”. For this task, we used MedGemma-27B \citep{sellergren2025medgemmatechnicalreport}, a state-of-the-art clinically specialized LLM.

\vspace{2cm}

\section{Author contributions}
Junmo Kim and Namkyeong Lee contributed to the conceptualization and design of the study. Junmo Kim handled and mainly analyzed the research data, and all authors interpreted the results. Junmo Kim organized deep learning models and LLM implementation. Junmo Kim wrote the original draft of the paper. Junmo Kim and Jiwon Kim made figures. Kwangsoo Kim supervised the study and provided computational resources. All authors had the final responsibility to submit for publication.

\section{Author contributions}
This research was supported and funded by SNUH Lee Kun-hee Child Cancer \& Rare Disease Project, Republic of Korea (grant number : 22A-000-0000).

\section{Competing interests}
All authors declare no competing interests.

\section{Data Availability}
All data used in this study is publicly available. The data generated and analyzed during the study are available from the corresponding author upon reasonable request. 

\section{Code Availability}
The codes for the study were based on our GitHub repository (https://github.com/kicarussays/MedRep).  

\newpage
\onecolumn
\section{Supplementary Materials}

\begin{table}[h]
\centering
\caption*{Supplementary Table 1. Downstream task performance. All downstream tasks were performed with three random seeds and all values are shown as mean ± SD.}
\begin{tabular}{llllllllll}
\toprule
\textbf{Representation} & \textbf{MT} & \textbf{LLOS} & \textbf{RA} & \textbf{Pheno} & \textbf{UTI} & \textbf{Fx} & \textbf{Sepsis} & \textbf{PNA} & \textbf{MI} \\
\midrule
\multicolumn{10}{l}{\textit{Metric: AUROC}} \\
\multicolumn{10}{l}{\textit{Internal Validation}} \\
None & \begin{tabular}[c]{@{}l@{}}0.857\\ ±0.030\end{tabular} & \begin{tabular}[c]{@{}l@{}}0.767\\ ±0.018\end{tabular} & \textbf{\begin{tabular}[c]{@{}l@{}}0.751\\ ±0.041\end{tabular}} & \begin{tabular}[c]{@{}l@{}}0.741\\ ±0.004\end{tabular} & \begin{tabular}[c]{@{}l@{}}0.836\\ ±0.010\end{tabular} & \begin{tabular}[c]{@{}l@{}}0.772\\ ±0.019\end{tabular} & \begin{tabular}[c]{@{}l@{}}0.791\\ ±0.005\end{tabular} & \begin{tabular}[c]{@{}l@{}}0.817\\ ±0.006\end{tabular} & \begin{tabular}[c]{@{}l@{}}0.789\\ ±0.011\end{tabular} \\
MedTok & \begin{tabular}[c]{@{}l@{}}0.894\\ ±0.007\end{tabular} & \begin{tabular}[c]{@{}l@{}}0.817\\ ±0.014\end{tabular} & \begin{tabular}[c]{@{}l@{}}0.700\\ ±0.048\end{tabular} & \begin{tabular}[c]{@{}l@{}}0.764\\ ±0.005\end{tabular} & \begin{tabular}[c]{@{}l@{}}0.857\\ ±0.007\end{tabular} & \begin{tabular}[c]{@{}l@{}}0.811\\ ±0.021\end{tabular} & \begin{tabular}[c]{@{}l@{}}0.828\\ ±0.017\end{tabular} & \begin{tabular}[c]{@{}l@{}}0.855\\ ±0.006\end{tabular} & \begin{tabular}[c]{@{}l@{}}0.844\\ ±0.003\end{tabular} \\
MedRep & \textbf{\begin{tabular}[c]{@{}l@{}}0.926\\ ±0.014\end{tabular}} & \textbf{\begin{tabular}[c]{@{}l@{}}0.845\\ ±0.004\end{tabular}} & \begin{tabular}[c]{@{}l@{}}0.747\\ ±0.045\end{tabular} & \textbf{\begin{tabular}[c]{@{}l@{}}0.766\\ ±0.002\end{tabular}} & \textbf{\begin{tabular}[c]{@{}l@{}}0.873\\ ±0.004\end{tabular}} & \textbf{\begin{tabular}[c]{@{}l@{}}0.857\\ ±0.007\end{tabular}} & \textbf{\begin{tabular}[c]{@{}l@{}}0.853\\ ±0.005\end{tabular}} & \textbf{\begin{tabular}[c]{@{}l@{}}0.871\\ ±0.005\end{tabular}} & \textbf{\begin{tabular}[c]{@{}l@{}}0.855\\ ±0.008\end{tabular}} \\
\midrule
\multicolumn{10}{l}{\textit{External Validation}} \\
None & \begin{tabular}[c]{@{}l@{}}0.730\\ ±0.015\end{tabular} & \begin{tabular}[c]{@{}l@{}}0.636\\ ±0.016\end{tabular} & \begin{tabular}[c]{@{}l@{}}0.607\\ ±0.006\end{tabular} & \begin{tabular}[c]{@{}l@{}}0.643\\ ±0.012\end{tabular} & \begin{tabular}[c]{@{}l@{}}0.795\\ ±0.022\end{tabular} & \begin{tabular}[c]{@{}l@{}}0.688\\ ±0.013\end{tabular} & \begin{tabular}[c]{@{}l@{}}0.777\\ ±0.024\end{tabular} & \begin{tabular}[c]{@{}l@{}}0.808\\ ±0.016\end{tabular} & \begin{tabular}[c]{@{}l@{}}0.705\\ ±0.019\end{tabular} \\
MedTok & \begin{tabular}[c]{@{}l@{}}0.803\\ ±0.021\end{tabular} & \begin{tabular}[c]{@{}l@{}}0.677\\ ±0.007\end{tabular} & \begin{tabular}[c]{@{}l@{}}0.580\\ ±0.038\end{tabular} & \begin{tabular}[c]{@{}l@{}}0.632\\ ±0.009\end{tabular} & \begin{tabular}[c]{@{}l@{}}0.811\\ ±0.002\end{tabular} & \begin{tabular}[c]{@{}l@{}}0.824\\ ±0.013\end{tabular} & \begin{tabular}[c]{@{}l@{}}0.827\\ ±0.013\end{tabular} & \begin{tabular}[c]{@{}l@{}}0.862\\ ±0.005\end{tabular} & \begin{tabular}[c]{@{}l@{}}0.783\\ ±0.005\end{tabular} \\
MedRep & \textbf{\begin{tabular}[c]{@{}l@{}}0.883\\ ±0.009\end{tabular}} & \textbf{\begin{tabular}[c]{@{}l@{}}0.753\\ ±0.016\end{tabular}} & \textbf{\begin{tabular}[c]{@{}l@{}}0.652\\ ±0.022\end{tabular}} & \textbf{\begin{tabular}[c]{@{}l@{}}0.664\\ ±0.011\end{tabular}} & \textbf{\begin{tabular}[c]{@{}l@{}}0.823\\ ±0.006\end{tabular}} & \textbf{\begin{tabular}[c]{@{}l@{}}0.879\\ ±0.015\end{tabular}} & \textbf{\begin{tabular}[c]{@{}l@{}}0.844\\ ±0.005\end{tabular}} & \textbf{\begin{tabular}[c]{@{}l@{}}0.865\\ ±0.003\end{tabular}} & \textbf{\begin{tabular}[c]{@{}l@{}}0.816\\ ±0.036\end{tabular}} \\
\midrule
\multicolumn{10}{l}{\textit{Metric: AUPRC}} \\
\multicolumn{10}{l}{\textit{Internal Validation}} \\
None & \begin{tabular}[c]{@{}l@{}}0.272\\ ±0.062\end{tabular} & \begin{tabular}[c]{@{}l@{}}0.558\\ ±0.034\end{tabular} & \begin{tabular}[c]{@{}l@{}}0.149\\ ±0.113\end{tabular} & \begin{tabular}[c]{@{}l@{}}0.297\\ ±0.009\end{tabular} & \begin{tabular}[c]{@{}l@{}}0.577\\ ±0.041\end{tabular} & \begin{tabular}[c]{@{}l@{}}0.184\\ ±0.018\end{tabular} & \begin{tabular}[c]{@{}l@{}}0.283\\ ±0.023\end{tabular} & \begin{tabular}[c]{@{}l@{}}0.403\\ ±0.028\end{tabular} & \begin{tabular}[c]{@{}l@{}}0.111\\ ±0.021\end{tabular} \\
MedTok & \begin{tabular}[c]{@{}l@{}}0.353\\ ±0.011\end{tabular} & \begin{tabular}[c]{@{}l@{}}0.647\\ ±0.017\end{tabular} & \begin{tabular}[c]{@{}l@{}}0.101\\ ±0.045\end{tabular} & \textbf{\begin{tabular}[c]{@{}l@{}}0.317\\ ±0.001\end{tabular}} & \begin{tabular}[c]{@{}l@{}}0.605\\ ±0.039\end{tabular} & \begin{tabular}[c]{@{}l@{}}0.377\\ ±0.006\end{tabular} & \begin{tabular}[c]{@{}l@{}}0.421\\ ±0.045\end{tabular} & \begin{tabular}[c]{@{}l@{}}0.524\\ ±0.011\end{tabular} & \begin{tabular}[c]{@{}l@{}}0.224\\ ±0.031\end{tabular} \\
MedRep & \textbf{\begin{tabular}[c]{@{}l@{}}0.443\\ ±0.027\end{tabular}} & \textbf{\begin{tabular}[c]{@{}l@{}}0.693\\ ±0.028\end{tabular}} & \textbf{\begin{tabular}[c]{@{}l@{}}0.191\\ ±0.107\end{tabular}} & \begin{tabular}[c]{@{}l@{}}0.308\\ ±0.007\end{tabular} & \textbf{\begin{tabular}[c]{@{}l@{}}0.642\\ ±0.033\end{tabular}} & \textbf{\begin{tabular}[c]{@{}l@{}}0.466\\ ±0.023\end{tabular}} & \textbf{\begin{tabular}[c]{@{}l@{}}0.510\\ ±0.021\end{tabular}} & \textbf{\begin{tabular}[c]{@{}l@{}}0.590\\ ±0.023\end{tabular}} & \textbf{\begin{tabular}[c]{@{}l@{}}0.273\\ ±0.046\end{tabular}} \\
\midrule
\multicolumn{10}{l}{\textit{External Validation}} \\
None & \begin{tabular}[c]{@{}l@{}}0.165\\ ±0.016\end{tabular} & \begin{tabular}[c]{@{}l@{}}0.395\\ ±0.007\end{tabular} & \begin{tabular}[c]{@{}l@{}}0.132\\ ±0.000\end{tabular} & \begin{tabular}[c]{@{}l@{}}0.243\\ ±0.011\end{tabular} & \begin{tabular}[c]{@{}l@{}}0.599\\ ±0.045\end{tabular} & \begin{tabular}[c]{@{}l@{}}0.155\\ ±0.005\end{tabular} & \begin{tabular}[c]{@{}l@{}}0.459\\ ±0.077\end{tabular} & \begin{tabular}[c]{@{}l@{}}0.501\\ ±0.038\end{tabular} & \begin{tabular}[c]{@{}l@{}}0.227\\ ±0.050\end{tabular} \\
MedTok & \begin{tabular}[c]{@{}l@{}}0.233\\ ±0.027\end{tabular} & \begin{tabular}[c]{@{}l@{}}0.438\\ ±0.012\end{tabular} & \begin{tabular}[c]{@{}l@{}}0.119\\ ±0.012\end{tabular} & \begin{tabular}[c]{@{}l@{}}0.237\\ ±0.005\end{tabular} & \begin{tabular}[c]{@{}l@{}}0.661\\ ±0.001\end{tabular} & \begin{tabular}[c]{@{}l@{}}0.612\\ ±0.042\end{tabular} & \begin{tabular}[c]{@{}l@{}}0.598\\ ±0.051\end{tabular} & \begin{tabular}[c]{@{}l@{}}0.747\\ ±0.014\end{tabular} & \begin{tabular}[c]{@{}l@{}}0.377\\ ±0.054\end{tabular} \\
MedRep & \textbf{\begin{tabular}[c]{@{}l@{}}0.445\\ ±0.004\end{tabular}} & \textbf{\begin{tabular}[c]{@{}l@{}}0.539\\ ±0.026\end{tabular}} & \textbf{\begin{tabular}[c]{@{}l@{}}0.158\\ ±0.014\end{tabular}} & \textbf{\begin{tabular}[c]{@{}l@{}}0.246\\ ±0.004\end{tabular}} & \textbf{\begin{tabular}[c]{@{}l@{}}0.664\\ ±0.003\end{tabular}} & \textbf{\begin{tabular}[c]{@{}l@{}}0.738\\ ±0.026\end{tabular}} & \textbf{\begin{tabular}[c]{@{}l@{}}0.710\\ ±0.035\end{tabular}} & \textbf{\begin{tabular}[c]{@{}l@{}}0.772\\ ±0.005\end{tabular}} & \textbf{\begin{tabular}[c]{@{}l@{}}0.472\\ ±0.145\end{tabular}} \\
\bottomrule
\end{tabular}
\end{table}

\begin{table}[h]
\centering
\caption*{Supplementary Table 2. Downstream task performance on Med-BERT. All downstream tasks were performed with three random seeds and all values are shown as mean ± SD.}
\begin{tabular}{llllllllll}
\toprule
\textbf{Representation} & \textbf{MT} & \textbf{LLOS} & \textbf{RA} & \textbf{Pheno} & \textbf{UTI} & \textbf{Fx} & \textbf{Sepsis} & \textbf{PNA} & \textbf{MI} \\
\midrule
\multicolumn{10}{l}{\textit{Metric: AUROC}} \\
\multicolumn{10}{l}{\textit{Internal Validation}} \\
None & \begin{tabular}[c]{@{}l@{}}0.828\\ ±0.040\end{tabular} & \begin{tabular}[c]{@{}l@{}}0.767\\ ±0.016\end{tabular} & \begin{tabular}[c]{@{}l@{}}0.662\\ ±0.094\end{tabular} & \begin{tabular}[c]{@{}l@{}}0.742\\ ±0.002\end{tabular} & \begin{tabular}[c]{@{}l@{}}0.820\\ ±0.010\end{tabular} & \begin{tabular}[c]{@{}l@{}}0.752\\ ±0.005\end{tabular} & \begin{tabular}[c]{@{}l@{}}0.783\\ ±0.001\end{tabular} & \begin{tabular}[c]{@{}l@{}}0.808\\ ±0.015\end{tabular} & \begin{tabular}[c]{@{}l@{}}0.773\\ ±0.037\end{tabular} \\
MedTok & \begin{tabular}[c]{@{}l@{}}0.878\\ ±0.008\end{tabular} & \textbf{\begin{tabular}[c]{@{}l@{}}0.807\\ ±0.011\end{tabular}} & \begin{tabular}[c]{@{}l@{}}0.722\\ ±0.070\end{tabular} & \begin{tabular}[c]{@{}l@{}}0.753\\ ±0.006\end{tabular} & \begin{tabular}[c]{@{}l@{}}0.849\\ ±0.009\end{tabular} & \begin{tabular}[c]{@{}l@{}}0.818\\ ±0.013\end{tabular} & \begin{tabular}[c]{@{}l@{}}0.814\\ ±0.005\end{tabular} & \begin{tabular}[c]{@{}l@{}}0.835\\ ±0.012\end{tabular} & \begin{tabular}[c]{@{}l@{}}0.842\\ ±0.023\end{tabular} \\
MedRep & \textbf{\begin{tabular}[c]{@{}l@{}}0.899\\ ±0.007\end{tabular}} & \begin{tabular}[c]{@{}l@{}}0.806\\ ±0.020\end{tabular} & \textbf{\begin{tabular}[c]{@{}l@{}}0.769\\ ±0.045\end{tabular}} & \textbf{\begin{tabular}[c]{@{}l@{}}0.762\\ ±0.003\end{tabular}} & \textbf{\begin{tabular}[c]{@{}l@{}}0.859\\ ±0.005\end{tabular}} & \textbf{\begin{tabular}[c]{@{}l@{}}0.842\\ ±0.017\end{tabular}} & \textbf{\begin{tabular}[c]{@{}l@{}}0.840\\ ±0.015\end{tabular}} & \textbf{\begin{tabular}[c]{@{}l@{}}0.852\\ ±0.001\end{tabular}} & \textbf{\begin{tabular}[c]{@{}l@{}}0.855\\ ±0.014\end{tabular}} \\
\midrule
\multicolumn{10}{l}{\textit{External Validation}} \\
None & \begin{tabular}[c]{@{}l@{}}0.637\\ ±0.072\end{tabular} & \begin{tabular}[c]{@{}l@{}}0.629\\ ±0.005\end{tabular} & \begin{tabular}[c]{@{}l@{}}0.569\\ ±0.013\end{tabular} & \begin{tabular}[c]{@{}l@{}}0.618\\ ±0.005\end{tabular} & \begin{tabular}[c]{@{}l@{}}0.814\\ ±0.006\end{tabular} & \begin{tabular}[c]{@{}l@{}}0.644\\ ±0.018\end{tabular} & \begin{tabular}[c]{@{}l@{}}0.785\\ ±0.038\end{tabular} & \begin{tabular}[c]{@{}l@{}}0.825\\ ±0.011\end{tabular} & \begin{tabular}[c]{@{}l@{}}0.678\\ ±0.040\end{tabular} \\
MedTok & \begin{tabular}[c]{@{}l@{}}0.806\\ ±0.007\end{tabular} & \begin{tabular}[c]{@{}l@{}}0.665\\ ±0.029\end{tabular} & \textbf{\begin{tabular}[c]{@{}l@{}}0.634\\ ±0.008\end{tabular}} & \textbf{\begin{tabular}[c]{@{}l@{}}0.653\\ ±0.023\end{tabular}} & \begin{tabular}[c]{@{}l@{}}0.811\\ ±0.004\end{tabular} & \begin{tabular}[c]{@{}l@{}}0.845\\ ±0.031\end{tabular} & \begin{tabular}[c]{@{}l@{}}0.773\\ ±0.010\end{tabular} & \begin{tabular}[c]{@{}l@{}}0.795\\ ±0.016\end{tabular} & \begin{tabular}[c]{@{}l@{}}0.783\\ ±0.014\end{tabular} \\
MedRep & \textbf{\begin{tabular}[c]{@{}l@{}}0.826\\ ±0.016\end{tabular}} & \textbf{\begin{tabular}[c]{@{}l@{}}0.713\\ ±0.026\end{tabular}} & \begin{tabular}[c]{@{}l@{}}0.621\\ ±0.004\end{tabular} & \begin{tabular}[c]{@{}l@{}}0.647\\ ±0.023\end{tabular} & \textbf{\begin{tabular}[c]{@{}l@{}}0.814\\ ±0.002\end{tabular}} & \textbf{\begin{tabular}[c]{@{}l@{}}0.861\\ ±0.010\end{tabular}} & \textbf{\begin{tabular}[c]{@{}l@{}}0.825\\ ±0.018\end{tabular}} & \textbf{\begin{tabular}[c]{@{}l@{}}0.866\\ ±0.006\end{tabular}} & \textbf{\begin{tabular}[c]{@{}l@{}}0.790\\ ±0.027\end{tabular}} \\
\midrule
\multicolumn{10}{l}{\textit{Metric: AUPRC}} \\
\multicolumn{10}{l}{\textit{Internal Validation}} \\
None & \begin{tabular}[c]{@{}l@{}}0.223\\ ±0.034\end{tabular} & \begin{tabular}[c]{@{}l@{}}0.562\\ ±0.021\end{tabular} & \begin{tabular}[c]{@{}l@{}}0.097\\ ±0.062\end{tabular} & \begin{tabular}[c]{@{}l@{}}0.293\\ ±0.005\end{tabular} & \begin{tabular}[c]{@{}l@{}}0.563\\ ±0.038\end{tabular} & \begin{tabular}[c]{@{}l@{}}0.173\\ ±0.013\end{tabular} & \begin{tabular}[c]{@{}l@{}}0.329\\ ±0.051\end{tabular} & \begin{tabular}[c]{@{}l@{}}0.426\\ ±0.023\end{tabular} & \begin{tabular}[c]{@{}l@{}}0.143\\ ±0.043\end{tabular} \\
MedTok & \begin{tabular}[c]{@{}l@{}}0.310\\ ±0.010\end{tabular} & \begin{tabular}[c]{@{}l@{}}0.631\\ ±0.019\end{tabular} & \textbf{\begin{tabular}[c]{@{}l@{}}0.152\\ ±0.133\end{tabular}} & \begin{tabular}[c]{@{}l@{}}0.304\\ ±0.005\end{tabular} & \begin{tabular}[c]{@{}l@{}}0.592\\ ±0.031\end{tabular} & \begin{tabular}[c]{@{}l@{}}0.322\\ ±0.014\end{tabular} & \begin{tabular}[c]{@{}l@{}}0.352\\ ±0.019\end{tabular} & \begin{tabular}[c]{@{}l@{}}0.422\\ ±0.013\end{tabular} & \begin{tabular}[c]{@{}l@{}}0.195\\ ±0.037\end{tabular} \\
MedRep & \textbf{\begin{tabular}[c]{@{}l@{}}0.381\\ ±0.040\end{tabular}} & \textbf{\begin{tabular}[c]{@{}l@{}}0.642\\ ±0.036\end{tabular}} & \begin{tabular}[c]{@{}l@{}}0.132\\ ±0.068\end{tabular} & \textbf{\begin{tabular}[c]{@{}l@{}}0.310\\ ±0.010\end{tabular}} & \textbf{\begin{tabular}[c]{@{}l@{}}0.612\\ ±0.035\end{tabular}} & \textbf{\begin{tabular}[c]{@{}l@{}}0.401\\ ±0.012\end{tabular}} & \textbf{\begin{tabular}[c]{@{}l@{}}0.471\\ ±0.044\end{tabular}} & \textbf{\begin{tabular}[c]{@{}l@{}}0.544\\ ±0.011\end{tabular}} & \textbf{\begin{tabular}[c]{@{}l@{}}0.215\\ ±0.002\end{tabular}} \\
\midrule
\multicolumn{10}{l}{\textit{External Validation}} \\
None & \begin{tabular}[c]{@{}l@{}}0.148\\ ±0.020\end{tabular} & \begin{tabular}[c]{@{}l@{}}0.371\\ ±0.007\end{tabular} & \begin{tabular}[c]{@{}l@{}}0.115\\ ±0.005\end{tabular} & \begin{tabular}[c]{@{}l@{}}0.224\\ ±0.008\end{tabular} & \begin{tabular}[c]{@{}l@{}}0.648\\ ±0.005\end{tabular} & \begin{tabular}[c]{@{}l@{}}0.148\\ ±0.006\end{tabular} & \begin{tabular}[c]{@{}l@{}}0.484\\ ±0.132\end{tabular} & \begin{tabular}[c]{@{}l@{}}0.629\\ ±0.034\end{tabular} & \begin{tabular}[c]{@{}l@{}}0.272\\ ±0.110\end{tabular} \\
MedTok & \begin{tabular}[c]{@{}l@{}}0.238\\ ±0.012\end{tabular} & \begin{tabular}[c]{@{}l@{}}0.427\\ ±0.020\end{tabular} & \begin{tabular}[c]{@{}l@{}}0.134\\ ±0.009\end{tabular} & \textbf{\begin{tabular}[c]{@{}l@{}}0.245\\ ±0.021\end{tabular}} & \begin{tabular}[c]{@{}l@{}}0.654\\ ±0.006\end{tabular} & \begin{tabular}[c]{@{}l@{}}0.624\\ ±0.073\end{tabular} & \begin{tabular}[c]{@{}l@{}}0.472\\ ±0.035\end{tabular} & \begin{tabular}[c]{@{}l@{}}0.516\\ ±0.036\end{tabular} & \textbf{\begin{tabular}[c]{@{}l@{}}0.359\\ ±0.063\end{tabular}} \\
MedRep & \textbf{\begin{tabular}[c]{@{}l@{}}0.342\\ ±0.032\end{tabular}} & \textbf{\begin{tabular}[c]{@{}l@{}}0.478\\ ±0.025\end{tabular}} & \textbf{\begin{tabular}[c]{@{}l@{}}0.137\\ ±0.001\end{tabular}} & \begin{tabular}[c]{@{}l@{}}0.243\\ ±0.015\end{tabular} & \textbf{\begin{tabular}[c]{@{}l@{}}0.654\\ ±0.009\end{tabular}} & \textbf{\begin{tabular}[c]{@{}l@{}}0.655\\ ±0.033\end{tabular}} & \textbf{\begin{tabular}[c]{@{}l@{}}0.658\\ ±0.055\end{tabular}} & \textbf{\begin{tabular}[c]{@{}l@{}}0.762\\ ±0.004\end{tabular}} & \begin{tabular}[c]{@{}l@{}}0.356\\ ±0.084\end{tabular} \\
\bottomrule
\end{tabular}
\end{table}

\begin{table}[h]
\centering
\caption*{Supplementary Table 3. Downstream task performance on CDM-BERT. All downstream tasks were performed with three random seeds and all values are shown as mean ± SD}
\begin{tabular}{llllllllll}
\toprule
\textbf{Representation} & \textbf{MT} & \textbf{LLOS} & \textbf{RA} & \textbf{Pheno} & \textbf{UTI} & \textbf{Fx} & \textbf{Sepsis} & \textbf{PNA} & \textbf{MI} \\
\midrule
\multicolumn{10}{l}{\textit{Metric: AUROC}} \\
\multicolumn{10}{l}{\textit{Internal Validation}} \\
None & \begin{tabular}[c]{@{}l@{}}0.862\\ ±0.010\end{tabular} & \begin{tabular}[c]{@{}l@{}}0.811\\ ±0.020\end{tabular} & \begin{tabular}[c]{@{}l@{}}0.730\\ ±0.062\end{tabular} & \begin{tabular}[c]{@{}l@{}}0.746\\ ±0.003\end{tabular} & \begin{tabular}[c]{@{}l@{}}0.852\\ ±0.005\end{tabular} & \begin{tabular}[c]{@{}l@{}}0.750\\ ±0.026\end{tabular} & \begin{tabular}[c]{@{}l@{}}0.791\\ ±0.042\end{tabular} & \begin{tabular}[c]{@{}l@{}}0.836\\ ±0.011\end{tabular} & \begin{tabular}[c]{@{}l@{}}0.730\\ ±0.059\end{tabular} \\
MedTok & \begin{tabular}[c]{@{}l@{}}0.879\\ ±0.012\end{tabular} & \begin{tabular}[c]{@{}l@{}}0.824\\ ±0.013\end{tabular} & \textbf{\begin{tabular}[c]{@{}l@{}}0.740\\ ±0.143\end{tabular}} & \begin{tabular}[c]{@{}l@{}}0.741\\ ±0.003\end{tabular} & \begin{tabular}[c]{@{}l@{}}0.855\\ ±0.010\end{tabular} & \begin{tabular}[c]{@{}l@{}}0.818\\ ±0.015\end{tabular} & \begin{tabular}[c]{@{}l@{}}0.835\\ ±0.010\end{tabular} & \begin{tabular}[c]{@{}l@{}}0.849\\ ±0.008\end{tabular} & \begin{tabular}[c]{@{}l@{}}0.843\\ ±0.011\end{tabular} \\
MedRep & \textbf{\begin{tabular}[c]{@{}l@{}}0.928\\ ±0.011\end{tabular}} & \textbf{\begin{tabular}[c]{@{}l@{}}0.858\\ ±0.006\end{tabular}} & \begin{tabular}[c]{@{}l@{}}0.733\\ ±0.033\end{tabular} & \textbf{\begin{tabular}[c]{@{}l@{}}0.766\\ ±0.007\end{tabular}} & \textbf{\begin{tabular}[c]{@{}l@{}}0.869\\ ±0.005\end{tabular}} & \textbf{\begin{tabular}[c]{@{}l@{}}0.837\\ ±0.017\end{tabular}} & \textbf{\begin{tabular}[c]{@{}l@{}}0.848\\ ±0.003\end{tabular}} & \textbf{\begin{tabular}[c]{@{}l@{}}0.871\\ ±0.007\end{tabular}} & \textbf{\begin{tabular}[c]{@{}l@{}}0.862\\ ±0.019\end{tabular}} \\
\midrule
\multicolumn{10}{l}{\textit{External Validation}} \\
None & \begin{tabular}[c]{@{}l@{}}0.770\\ ±0.004\end{tabular} & \begin{tabular}[c]{@{}l@{}}0.699\\ ±0.010\end{tabular} & \begin{tabular}[c]{@{}l@{}}0.597\\ ±0.025\end{tabular} & \begin{tabular}[c]{@{}l@{}}0.623\\ ±0.021\end{tabular} & \begin{tabular}[c]{@{}l@{}}0.810\\ ±0.002\end{tabular} & \begin{tabular}[c]{@{}l@{}}0.714\\ ±0.037\end{tabular} & \begin{tabular}[c]{@{}l@{}}0.844\\ ±0.007\end{tabular} & \begin{tabular}[c]{@{}l@{}}0.840\\ ±0.017\end{tabular} & \begin{tabular}[c]{@{}l@{}}0.703\\ ±0.054\end{tabular} \\
MedTok & \begin{tabular}[c]{@{}l@{}}0.842\\ ±0.015\end{tabular} & \begin{tabular}[c]{@{}l@{}}0.700\\ ±0.014\end{tabular} & \textbf{\begin{tabular}[c]{@{}l@{}}0.631\\ ±0.027\end{tabular}} & \begin{tabular}[c]{@{}l@{}}0.639\\ ±0.025\end{tabular} & \begin{tabular}[c]{@{}l@{}}0.800\\ ±0.006\end{tabular} & \begin{tabular}[c]{@{}l@{}}0.842\\ ±0.012\end{tabular} & \begin{tabular}[c]{@{}l@{}}0.849\\ ±0.005\end{tabular} & \begin{tabular}[c]{@{}l@{}}0.855\\ ±0.003\end{tabular} & \begin{tabular}[c]{@{}l@{}}0.802\\ ±0.011\end{tabular} \\
MedRep & \textbf{\begin{tabular}[c]{@{}l@{}}0.876\\ ±0.013\end{tabular}} & \textbf{\begin{tabular}[c]{@{}l@{}}0.764\\ ±0.008\end{tabular}} & \begin{tabular}[c]{@{}l@{}}0.626\\ ±0.023\end{tabular} & \textbf{\begin{tabular}[c]{@{}l@{}}0.645\\ ±0.009\end{tabular}} & \textbf{\begin{tabular}[c]{@{}l@{}}0.819\\ ±0.005\end{tabular}} & \textbf{\begin{tabular}[c]{@{}l@{}}0.873\\ ±0.003\end{tabular}} & \textbf{\begin{tabular}[c]{@{}l@{}}0.859\\ ±0.007\end{tabular}} & \textbf{\begin{tabular}[c]{@{}l@{}}0.865\\ ±0.014\end{tabular}} & \textbf{\begin{tabular}[c]{@{}l@{}}0.826\\ ±0.007\end{tabular}} \\
\midrule
\multicolumn{10}{l}{\textit{Metric: AUPRC}} \\
\multicolumn{10}{l}{\textit{Internal Validation}} \\
None & \begin{tabular}[c]{@{}l@{}}0.318\\ ±0.076\end{tabular} & \begin{tabular}[c]{@{}l@{}}0.621\\ ±0.037\end{tabular} & \begin{tabular}[c]{@{}l@{}}0.130\\ ±0.047\end{tabular} & \begin{tabular}[c]{@{}l@{}}0.291\\ ±0.004\end{tabular} & \begin{tabular}[c]{@{}l@{}}0.606\\ ±0.041\end{tabular} & \begin{tabular}[c]{@{}l@{}}0.235\\ ±0.011\end{tabular} & \begin{tabular}[c]{@{}l@{}}0.415\\ ±0.006\end{tabular} & \begin{tabular}[c]{@{}l@{}}0.471\\ ±0.038\end{tabular} & \begin{tabular}[c]{@{}l@{}}0.194\\ ±0.060\end{tabular} \\
MedTok & \begin{tabular}[c]{@{}l@{}}0.394\\ ±0.015\end{tabular} & \begin{tabular}[c]{@{}l@{}}0.669\\ ±0.018\end{tabular} & \textbf{\begin{tabular}[c]{@{}l@{}}0.136\\ ±0.110\end{tabular}} & \begin{tabular}[c]{@{}l@{}}0.290\\ ±0.000\end{tabular} & \begin{tabular}[c]{@{}l@{}}0.623\\ ±0.042\end{tabular} & \textbf{\begin{tabular}[c]{@{}l@{}}0.427\\ ±0.030\end{tabular}} & \begin{tabular}[c]{@{}l@{}}0.504\\ ±0.009\end{tabular} & \begin{tabular}[c]{@{}l@{}}0.546\\ ±0.007\end{tabular} & \begin{tabular}[c]{@{}l@{}}0.256\\ ±0.019\end{tabular} \\
MedRep & \textbf{\begin{tabular}[c]{@{}l@{}}0.438\\ ±0.021\end{tabular}} & \textbf{\begin{tabular}[c]{@{}l@{}}0.715\\ ±0.019\end{tabular}} & \begin{tabular}[c]{@{}l@{}}0.096\\ ±0.042\end{tabular} & \textbf{\begin{tabular}[c]{@{}l@{}}0.312\\ ±0.004\end{tabular}} & \textbf{\begin{tabular}[c]{@{}l@{}}0.626\\ ±0.032\end{tabular}} & \begin{tabular}[c]{@{}l@{}}0.413\\ ±0.011\end{tabular} & \textbf{\begin{tabular}[c]{@{}l@{}}0.514\\ ±0.025\end{tabular}} & \textbf{\begin{tabular}[c]{@{}l@{}}0.598\\ ±0.014\end{tabular}} & \textbf{\begin{tabular}[c]{@{}l@{}}0.328\\ ±0.044\end{tabular}} \\
\midrule
\multicolumn{10}{l}{\textit{External Validation}} \\
None & \begin{tabular}[c]{@{}l@{}}0.226\\ ±0.018\end{tabular} & \begin{tabular}[c]{@{}l@{}}0.482\\ ±0.009\end{tabular} & \begin{tabular}[c]{@{}l@{}}0.118\\ ±0.010\end{tabular} & \begin{tabular}[c]{@{}l@{}}0.225\\ ±0.019\end{tabular} & \begin{tabular}[c]{@{}l@{}}0.649\\ ±0.004\end{tabular} & \begin{tabular}[c]{@{}l@{}}0.239\\ ±0.005\end{tabular} & \begin{tabular}[c]{@{}l@{}}0.720\\ ±0.012\end{tabular} & \begin{tabular}[c]{@{}l@{}}0.664\\ ±0.061\end{tabular} & \begin{tabular}[c]{@{}l@{}}0.364\\ ±0.029\end{tabular} \\
MedTok & \begin{tabular}[c]{@{}l@{}}0.324\\ ±0.020\end{tabular} & \begin{tabular}[c]{@{}l@{}}0.482\\ ±0.015\end{tabular} & \begin{tabular}[c]{@{}l@{}}0.126\\ ±0.014\end{tabular} & \begin{tabular}[c]{@{}l@{}}0.234\\ ±0.020\end{tabular} & \begin{tabular}[c]{@{}l@{}}0.643\\ ±0.003\end{tabular} & \begin{tabular}[c]{@{}l@{}}0.694\\ ±0.019\end{tabular} & \begin{tabular}[c]{@{}l@{}}0.749\\ ±0.003\end{tabular} & \begin{tabular}[c]{@{}l@{}}0.768\\ ±0.005\end{tabular} & \begin{tabular}[c]{@{}l@{}}0.508\\ ±0.017\end{tabular} \\
MedRep & \textbf{\begin{tabular}[c]{@{}l@{}}0.400\\ ±0.040\end{tabular}} & \textbf{\begin{tabular}[c]{@{}l@{}}0.563\\ ±0.012\end{tabular}} & \textbf{\begin{tabular}[c]{@{}l@{}}0.136\\ ±0.018\end{tabular}} & \textbf{\begin{tabular}[c]{@{}l@{}}0.243\\ ±0.009\end{tabular}} & \textbf{\begin{tabular}[c]{@{}l@{}}0.663\\ ±0.001\end{tabular}} & \textbf{\begin{tabular}[c]{@{}l@{}}0.720\\ ±0.003\end{tabular}} & \textbf{\begin{tabular}[c]{@{}l@{}}0.750\\ ±0.016\end{tabular}} & \textbf{\begin{tabular}[c]{@{}l@{}}0.772\\ ±0.015\end{tabular}} & \textbf{\begin{tabular}[c]{@{}l@{}}0.552\\ ±0.008\end{tabular}} \\
\bottomrule
\end{tabular}
\end{table}

\begin{table}[h]
\centering
\caption*{Supplementary Table 4. Medical vocabularies included in MedRep. }
\begin{tabular}{ll|ll}
\toprule
Vocabulary & Count & Vocabulary & Count \\
\midrule
RxNorm Extension & 2145325 & CIM10 & 14223 \\
NDC & 1183702 & ICD9ProcCN & 13385 \\
SPL & 720063 & CCAM & 10206 \\
SNOMED & 531174 & OPCS4 & 9821 \\
RxNorm & 308709 & Multum & 9770 \\
EDI & 293618 & CTD & 8698 \\
OMOP Genomic & 289889 & ClinVar & 8072 \\
Nebraska Lexicon & 268995 & JAX & 7855 \\
ICD10PCS & 196221 & OXMIS & 7083 \\
dm+d & 192706 & ATC & 6740 \\
DPD & 188287 & Cancer Modifier & 6028 \\
LOINC & 155903 & HemOnc & 5786 \\
AMT & 130006 & OncoKB & 5569 \\
ICD10CM & 88467 & CGI & 5351 \\
Read & 74816 & ICD9Proc & 4657 \\
NDFRT & 69566 & SNOMED Veterinary & 4009 \\
KDC & 63749 & SUS & 3914 \\
ICDO3 & 62535 & HCPCS & 3422 \\
NCCD & 51547 & UK Biobank & 2725 \\
CIEL & 48544 & NCIt & 2426 \\
BDPM & 43214 & NAACCR & 2173 \\
OPS & 42959 & CIViC & 1386 \\
JMDC & 37485 & OMOP Extension & 1189 \\
ICD10CN & 31309 & EphMRA ATC & 895 \\
OMOP Invest Drug & 29727 & OncoTree & 885 \\
GCN\_SEQNO & 29659 & NFC & 692 \\
VANDF & 29351 & VA Class & 576 \\
GGR & 25849 & CO-CONNECT TWINS & 449 \\
MeSH & 24376 & SMQ & 324 \\
KCD7 & 20010 & CVX & 271 \\
ICD10GM & 16250 & CO-CONNECT & 137 \\
ICD9CM & 15585 & PPI & 109 \\
ICD10 & 14411 & Cohort & 78 \\
\bottomrule

\end{tabular}
\end{table}

\begin{table}[t]
\centering
{\footnotesize
  \caption*{Supplementary Table 5. Prompt for each domain of OMOP vocabulary and example descriptions.}
  \vspace{0.5cm}
  {
\begin{tabular}{ p{0.08\linewidth}  p{0.30\linewidth}  p{0.45\linewidth} }
  \toprule
  \textbf{Domain} & \textbf{Prompt (with example concept)} & \textbf{Description}  \\
  \hline
  Condition & 
  Instruction: Briefly explain the clinical background and regarding treatments of each concept name (condition) with less than 5 sentences. Do not include sentences that are too ordinary (such as "further details would depend on the specific situation) and focus on describing the representative clinical features of the concept. Concept name: Neutropenic fever & 
  Neutropenic fever occurs in patients with neutropenia, characterized by an abnormally low level of neutrophils, often due to chemotherapy, bone marrow disorders, or immunosuppression. Clinically, it presents as a fever, which can be the only sign of infection in these patients due to their compromised immune response. The absence of typical signs of infection, such as pus formation or localized symptoms, can make diagnosis challenging. Treatment involves prompt administration of broad-spectrum antibiotics to mitigate the risk of severe infections, and in some cases, granulocyte-colony stimulating factors (G-CSF) may be used to expedite neutrophil recovery. Monitoring for potential complications, such as sepsis, is crucial in the management of neutropenic fever. \\
  \hline
  Drug & 
 Instruction: Briefly explain the clinical background and purpose of each concept name (drug) with less than 5 sentences. Do not include sentences that are too ordinary (such as "further details would depend on the specific situation) and focus on describing the representative clinical features of the concept. For explanation, if it exists in the concept name, take into account the detailed items of the concept such as ingredient, dosage form, and strength. If several drugs are contained in a concept, do not explain those drugs separately, but explain the concept name comprehensively and finish the answer with less than 5 sentences. Concept name: Aspirin 100 MG Oral Tablet [Aspirin] & 
  Aspirin 100 MG Oral Tablet is a low-dose formulation of acetylsalicylic acid, primarily used for its antiplatelet effects to reduce the risk of cardiovascular events such as heart attacks and strokes. This dosage is particularly effective in the long-term management of patients with a history of cardiovascular disease or those at high risk. Additionally, aspirin has anti-inflammatory and analgesic properties, making it useful in managing certain inflammatory conditions and mild to moderate pain. The 100 MG strength is often chosen to minimize gastrointestinal side effects while providing adequate antithrombotic benefits. Regular use under a healthcare provider’s guidance is crucial for balancing benefits and potential risks. \\
  \hline
  Measurement & 
  Instruction: Briefly explain the clinical background and context of each concept name (measurement) with less than 5 sentences. Do not include sentences that are too ordinary (such as "further details would depend on the specific situation) and focus on describing the representative clinical features of the concept. For explanation, if it exists in the concept name, describe what the decile means clinically. Concept name: Glycated haemoglobin (HbA1c) (9th decile) & 
  Glycated hemoglobin (HbA1c) is a measure of average blood glucose levels over the past two to three months, reflecting long-term glycemic control in individuals with diabetes. A value in the 9th decile typically indicates a HbA1c level of approximately 8.0\% to 9.0\%, suggesting poorly controlled diabetes and an increased risk for microvascular and macrovascular complications, such as retinopathy and cardiovascular disease. Clinically, this decile serves as a critical threshold for considering more aggressive therapeutic interventions, including medication adjustments and comprehensive diabetes management strategies. Monitoring HbA1c in this range underscores the importance of lifestyle modifications and personalized treatment plans to improve glycemic control and reduce long-term health risks. \\
  \hline
  Procedure & 
  Instruction: Briefly explain the clinical background and purpose of each concept name (procedure) with less than 5 sentences. Do not include sentences that are too ordinary (such as "further details would depend on the specific situation) and focus on describing the representative clinical features of the concept. Concept name: Holter extended electrocardiographic recording & 
  Holter extended electrocardiographic recording is a continuous monitoring technique that captures the heart's electrical activity over 24 to 48 hours using a portable ECG device. This procedure is essential for diagnosing arrhythmias, assessing symptoms like palpitations, and evaluating the efficacy of antiarrhythmic medications. Patients wear the device during their daily activities, allowing for a comprehensive assessment of heart rhythm in real-life settings rather than just during a brief clinical visit. Analysis of the recorded data can reveal transient abnormalities or patterns that might not be evident during a standard ECG. \\
  \bottomrule
    \end{tabular}}}
\end{table}

\begin{table}[h]
\centering
\caption*{Supplementary Table 6. Configuration for GRACE}
\begin{tabular}{ll}
\toprule
\textbf{Configuration} & \textbf{Value} \\
\midrule
Batch size & 1024 \\
Learning rate & 0.0005 \\
Hidden size & 768 \\
Activation function & pReLU \\
Graph neural network & Graph convolutional networks \\
Number of layers & 2 \\
Dropout rate & 0.2 \\
DropEdge rate & 0.2 \\
$\tau$ & 0.5 \\
Max epoch & 200 \\
Early stopping patience & 20 \\
\bottomrule
\end{tabular}
\end{table}

\end{document}